%% file: main.tex
\documentclass[preprint,12pt]{elsarticle}

\usepackage{amssymb}
\usepackage{amsmath}

\usepackage{graphicx}
\usepackage{algorithmic}
\usepackage[ruled]{algorithm2e}
\usepackage{array}
\usepackage[caption=false,font=normalsize,labelfont=sf,textfont=sf]{subfig}
\usepackage{textcomp}
\usepackage{stfloats}
\usepackage{url}
\usepackage{float}

\usepackage{hyperref}
\hypersetup{
    breaklinks=true,
    colorlinks=true,
    urlcolor=blue,
    citecolor=blue,
    linkcolor=blue
}

\usepackage{verbatim}

\usepackage{booktabs}
\usepackage{color}
\usepackage[dvipsnames]{xcolor}
\newcommand{\cmark}{\text{\checkmark}}  
\newcommand{\xmark}{\text{\ding{55}}}   
\usepackage{pifont}  
\usepackage{xr-hyper}

\usepackage{multirow}

\def\BibTeX{{\rm B\kern-.05em{\sc i\kern-.025em b}\kern-.08em
    T\kern-.1667em\lower.7ex\hbox{E}\kern-.125emX}}

\journal{Nuclear Physics B}

\begin{document}

\begin{frontmatter}



\title{I-Segmenter: Integer-Only Vision Transformer for Efficient Semantic Segmentation}


\author[1]{Jordan Sassoon}
\ead{jordan.sassoon@cea.fr}

\author[1]{Michal Szczepanski}
\ead{michal.szczepanski@cea.fr}

\author[1]{Martyna Poreba}
\ead{martyna.poreba@cea.fr}

\affiliation[1]{
    organization={Université Paris-Saclay, CEA, List},
    city={Palaiseau},
    postcode={F-91120},
    country={France}
}



\begin{abstract}
Vision Transformers (ViTs) have recently achieved strong results in semantic segmentation, yet their deployment on resource-constrained devices remains limited due to their high memory footprint and computational cost. Quantization offers an effective strategy to improve efficiency, but ViT-based segmentation models are notoriously fragile under low precision, as quantization errors accumulate across deep encoder–decoder pipelines. We introduce I-Segmenter, the first fully integer-only ViT segmentation framework. Building on the Segmenter architecture, I-Segmenter systematically replaces floating-point operations with integer-only counterparts. To further stabilize both training and inference, we propose $\lambda$-ShiftGELU, a novel activation function that mitigates the limitations of uniform quantization in handling long-tailed activation distributions. In addition, we remove the L2 normalization layer and replace bilinear interpolation in the decoder with nearest-neighbor upsampling, ensuring integer-only execution throughout the computational graph. Extensive experiments show that I-Segmenter achieves accuracy within a reasonable margin of its FP32 baseline (\(5.1\%\) on average), while reducing model size by up to \(3.8\times\) and enabling up to \(1.2\times\) faster inference with optimized runtimes. Notably, even in one-shot PTQ with a single calibration image, I-Segmenter delivers competitive accuracy, underscoring its practicality for real-world deployment. Code is available at https://github.com/ms245755/I-segmenter.
\end{abstract}

\begin{keyword}
ViT, Semantic Segmentation, Quantization, Integer. 


\end{keyword}

\end{frontmatter}


\input{1_intro}  
\input{2_sota}  
\input{3_methodology}  
\input{4_experimental_setup}  
\input{5_ablation}  
\input{6_results}  
\input{7_conclusion}  



\appendix

\input{A1_pseudocode}
\newpage
\input{A2_prediction_comparison}

\clearpage

\bibliographystyle{elsarticle-num} 


\input main.bbl


\end{document}

%% file: 1_intro.tex
\section{Introduction}
\label{sec:Intro}
By enabling global context modeling, Vision Transformers (ViTs) have pushed semantic segmentation to new state-of-the-art results \cite{SETR, Zhang_2024, Ranftl2021, strudel2021segmenter,xie2021segformersimpleefficientdesign, zhang2022segvitsemanticsegmentationplain, zhang2023segvitv2exploringefficientcontinual, StructToken, cheng2021perpixelclassificationneedsemantic, cheng2021mask2former, kerssies2025eomt}. These modern frameworks typically adopt ViTs as encoders combined with either CNN-based or Transformer-based \cite{vaswani2017attention} decoders, surpassing the precision of fully CNN-based architectures. Achieving accurate segmentation, however, requires preserving fine local details across high-resolution inputs. Meeting the need for dense patch-level information places a heavy burden on ViTs, as the self-attention mechanism scales quadratically with input size, causing sharp increases in memory and computation. To cope with this complexity, several research directions have been investigated \cite{bondarenko2021understandingovercomingchallengesefficient, chittyvenkata2023surveytechniquesoptimizingtransformer, xu2024comprehensive, SAHA2025130417}. Compact architectures \cite{mehta2022mobilevitlightweightgeneralpurposemobilefriendly,li2023rethinkingvisiontransformersmobilenet,pan2022edgevitscompetinglightweightcnns, chen2022mobileformerbridgingmobilenettransformer, MicoViT25, Dong2025LightViT} redesign Transformer components to balance accuracy with efficiency. Model compression techniques such as pruning \cite{norouzi2024algm, proust2025step, PM-ViT, CHEN2025128747, 10880106, MARCHETTI2025127449,10483924}, quantization \cite{wu2024adalog, tang2024easyquant, xiao2024smoothquant, yuan2024ptq4vit, kim2025mixednonlinear, wu2025fimaq, HE2025103530}, and knowledge distillation \cite{touvron2021trainingdataefficientimagetransformers,habib2024knowledgedistillationvisiontransformers} aim to reduce computational cost without severely impacting accuracy.
Quantization reduces numerical precision by representing values and operations with fewer bits, mapping them from high-precision formats to lower-precision ones. This yields faster, smaller, and more memory-efficient models.
Integer quantization offers substantial advantages for model deployment. By converting floating-point representations (FP) into compact integer formats (INT), it reduces memory footprint and bandwidth requirements, accelerates inference on hardware optimized for integer arithmetic, and lowers energy consumption. Such gains are crucial for edge deployment on constrained hardware, including ARM Cortex-M microcontrollers, Qualcomm Snapdragon CPUs, and dedicated NPUs such as Google Edge TPU or ARM Ethos-U. Two main strategies dominate: post-training quantization (PTQ) and quantization-aware training (QAT). PTQ converts a pretrained model to low precision using a small calibration set, enabling fast deployment but frequently degrading accuracy. In contrast, QAT simulates quantization effects during training, allowing the model to adapt its weights to low-precision constraints, often achieving higher accuracy at the cost of increased training complexity. In general, 8- or 6-bit quantization results in about a 3\% drop in Top-1 accuracy on ImageNet-1k \cite{5206848}. At 4 bits, this degradation becomes much more pronounced, with early methods showing losses of up to 40\%. At 3 bits, the loss is even more severe, with accuracy collapsing dramatically and some models becoming nearly unusable \cite{wu2025fimaq}. A central challenge in ViT quantization comes from non-linear operations. Small numerical errors can be disproportionately amplified. In Softmax, tiny logit perturbations cause large shifts in attention, distorting token relationships. GELU adds further difficulty, its smooth non-linearity and long-tailed activations make it hard to preserve both small values and outliers under low precision.

Although quantization has shown promising results for ViTs in image classification, its application to semantic segmentation remains largely unexplored. Segmentation is particularly sensitive to quantization errors, as it requires dense pixel-level predictions. While individual ViT components can tolerate reduced precision, errors accumulate across the many encoder blocks, making the model highly vulnerable under low-bit quantization. The addition of a decoder head further complicates the problem, since multi-scale aggregation and nonlinear operations can amplify errors propagated from the backbone. In this work, we present the first systematic study of quantizing ViT-based models for semantic segmentation. Our contributions are the following:

\begin{itemize}
    \item We introduce I-Segmenter, a ViT-based semantic segmentation framework enabling end-to-end integer-only inference for dense prediction.
    
    \item We propose \(\lambda\)-ShiftGELU, an integer-compatible activation that improves the robustness of both QAT and PTQ.
    
    \item We revisit and adapt key architectural components of ViT-based segmentation model to integer-only execution, and analyze their impact through ablation studies.
    
    \item Extensive evaluations show that I-Segmenter preserves accuracy comparable to FP32 models under reduced numerical precision, including in low-calibration post-training quantization settings.
\end{itemize}

This paper is organized as follows. We review related work on ViT-based semantic segmentation and quantization.
We then introduce I-Segmenter and its integer-only design. Experimental settings and results, including ablation studies, are presented next. The paper concludes with a discussion.

%% file: 2_sota.tex
\section{Background and Related Work}
\label{sec:SOTA}
ViTs process an image as a sequence of fixed-size patches, each embedded into a vector representation. In their original formulation, a learnable $[CLS]$ token is prepended to this sequence to aggregate global information. These tokens are then processed by a stack of Transformer encoder layers, each composed of a Multi-Head Self-Attention (MHSA) module, followed by a feed-forward network (FFN). In practice, the MHSA computes pairwise interactions between all tokens by projecting them into query ($Q$), key ($K$), and value ($V$) representations, enabling the model to capture global dependencies. The FFN, also referred to as a multilayer perceptron (MLP), consists of stacked linear layers with a non-linear activation and is applied independently to each token.

\subsection{Segmentation Meets Vision Transformers}

ViTs were originally designed for image classification, where a single global representation, extracted from the $[CLS]$ token, is used by a MLP head to output one class label for the entire image. In contrast, semantic segmentation requires predicting a class label for each pixel, resulting in dense spatial outputs. This fundamental difference makes segmentation significantly more sensitive to numerical errors, as inaccuracies accumulate across all pixels and propagate through both the encoder and decoder. Moreover, these dense predictions increase the computational burden of the model, as fine-grained spatial information must be processed throughout the network.

Different ViT-based models achieve this objective through distinct decoding strategies. 
SETR \cite{SETR,Zhang_2024} was a pioneering effort that demonstrated the viability of pure Transformer models for dense prediction tasks. It adopts a ViT backbone as an encoder, processing an input image as a sequence of non-overlapping patches. To generate dense per-pixel class predictions, it uses a classic CNN-style upsampling decoder. Several variants depending on the decoder part of the model have been proposed, either performing direct bilinear upsampling (SETR-Naïve), applying progressive upsampling with convolutional blocks (SETR-PUP), or aggregating features from multiple Transformer layers to enhance spatial detail (SETR-MLA). While SETR showed that transformers could replace convolutional encoders in segmentation, it suffers from high memory and computational cost, mainly due to the full self-attention mechanism applied over all tokens. DPT \cite{Ranftl2021}, originally introduced for monocular depth estimation, is a versatile Transformer-based architecture that employs a ViT backbone and a multi-scale fusion decoder that preserves spatial details while benefiting from the global context modeling of transformers. The decoder in DPT is a lightweight module of three convolutional layers that converts high-resolution features into final predictions. It outputs class logits for segmentation or a depth map for regression, without batch normalization. Upsampling to the input resolution is done via strided convolutions or bilinear interpolation. StructToken \cite{StructToken} introduces learnable structure tokens that interact iteratively with patch representations. These structure-aware tokens gradually build segmentation masks while assigning corresponding class labels. Segmenter \cite{strudel2021segmenter} improves architectural modularity and coherence by proposing a fully Transformer-based encoder-decoder framework. Instead of a convolutional head, it introduces a set of learnable class-specific mask embeddings that attend to the encoder tokens, directly producing segmentation masks via attention. SegFormer \cite{xie2021segformersimpleefficientdesign} combines a hierarchical Transformer encoder with a lightweight MLP decoder for efficient semantic segmentation. The encoder, called Mix Transformer, extracts multi-scale features without positional embeddings, using overlapping patch merging and efficient self-attention to reduce computational cost. The decoder fuses these features using only MLP layers, avoiding complex modules like convolutions. This design enables SegFormer to achieve high accuracy with significantly lower memory and compute requirements than models like SETR. Unlike previous frameworks that generally adopt a per-pixel classification paradigm, SegViT \cite{zhang2022segvitsemanticsegmentationplain} and SegViT2 \cite{zhang2023segvitv2exploringefficientcontinual} introduce a different decoding philosophy centered on the attention mechanism itself. Its key contribution is the Attention-to-Mask (ATM) module, which directly converts similarity maps between learnable class tokens and spatial features into segmentation masks. This departs from prior designs that relied on convolutional upsampling (SETR), lightweight convolutional heads (DPT), or class embeddings refined through self-attention (Segmenter, StructToken). Both SegViT and SegViT2 adopt more complex designs but deliver improved segmentation quality, especially with large backbones, outperforming earlier models. Mask2Former \cite{cheng2021mask2former} builds on MaskFormer \cite{cheng2021perpixelclassificationneedsemantic} by enhancing both its decoder and multi-scale capability. It uses masked attention in the Transformer decoder and feeds feature maps of different resolutions. Together, the pixel decoder extracts detailed multi-scale representations, while the Transformer decoder, equipped with masked attention and scale-aware processing, produces precise masks and their class labels in an integrated pipeline. While Mask2Former excels at accuracy, its decoder complexity makes it onerous for resource-constrained environments. Recently, EoMT \cite{kerssies2025eomt} introduces a radically simplified vision of segmentation, demonstrating that a plain ViT can be repurposed to jointly encode both image patches and segmentation queries as tokens, without the need for adapters, pixel decoders, or Transformer decoders. This encoder-only solution strikes a markedly better balance between inference speed and segmentation quality, though it remains relatively computationally demanding compared to lightweight alternatives like SegViT or Segmenter.

\subsection{Challenges in Quantizing ViT}\label{Quant_Challenges}

\begin{figure*}[!t]
\centering
\includegraphics[width=0.98\linewidth]{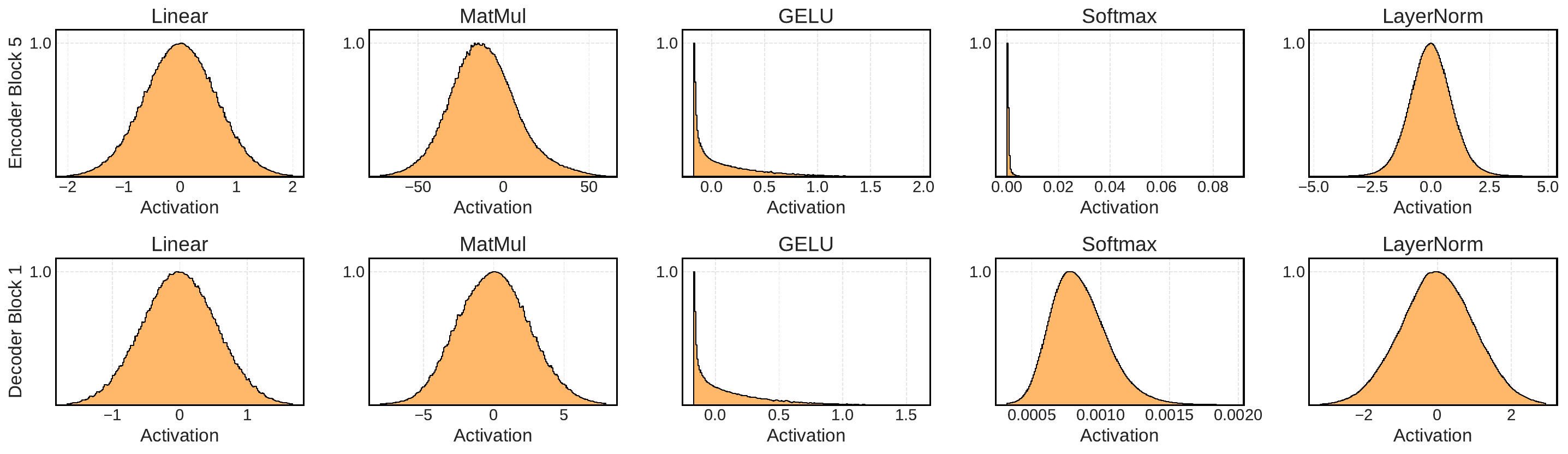}
\caption{Visualization of activation patterns across different layer types in the fifth encoder block and first decoder block of the Segmenter architecture. Here, the encoder corresponds to the ViT backbone and the decoder to the Mask Transformer head, with blocks counted sequentially within each component (e.g., the encoder typically contains 12–24 layers depending on the backbone, while the decoder consists of 2 layers). Linear, MatMul, and LayerNorm exhibit approximately Gaussian-like distributions, whereas GELU and Softmax produce highly skewed or long-tailed activations, making them significantly more challenging for uniform quantization.}
\label{fig:Op_distribution}
\end{figure*}

ViT quantization poses inherent challenges due to the architectural diversity of its operators. Components such as linear projections (Linear), matrix multiplications (MatMul), normalization layers (LayerNorm), activation functions (GELU), and attention mechanisms (Softmax) exhibit distinct distributional shapes and sensitivities to precision (Figure~\ref{fig:Op_distribution}). Quantization methods must accommodate both highly skewed distributions, such as those from activation functions, and more Gaussian-like distributions, such as those from linear projections. This heterogeneity makes it difficult to preserve accuracy across the network, particularly when using naive uniform quantization.

One of the central difficulties in quantizing ViT arises from the sensitivity of attention layers. MHSA relies on precise dot-product computations between query $Q$ and key $K$ vectors, which determine attention weights after normalization through the Softmax function. These $QK$ dot products are highly sensitive to precision loss, as small errors are amplified by the Softmax function, whose exponential mapping magnifies even minor quantization errors, ultimately leading to unstable and biased attention weights.

Reducing precision can therefore severely degrade attention scores. In addition, ViTs heavily rely on LayerNorm, whose non-linearity poses challenges for quantization since it prevents straightforward use of scaling factors or integer arithmetic through the dyadic assumption. LayerNorm' fragility is due to its dependence on small variances, making it highly susceptible to activation explosion under scaling errors. Another critical difficulty arises from activation outliers: both logits and GELU activations exhibit long-tailed distributions. Uniform quantization allocates bins uniformly across the entire range, which wastes resolution on rarely occurring large values while undersampling the dense region around the peak. Consequently, near-zero activations are not captured with sufficient precision, and rare but large activations are quantized too coarsely, introducing distortions that accumulate across layers and can severely impact model stability. This effect is exacerbated by the repeated stacking of Transformer blocks, where quantization errors accumulate and amplify with depth. Residual connections further amplify this issue, acting as direct linear pathways that propagate noise without attenuation and limiting the potential for error correction. Finally, positional encodings, whether sinusoidal or learned, are also vulnerable to quantization, as distortions risk distorting the spatial relationships essential for retaining the model's accuracy. Similarly, the class embedding is particularly sensitive, as it serves as the global information aggregator. Quantization noise on this token propagates to all subsequent layers, potentially degrading the quality of the final representation and reducing classification accuracy.

In classification settings, the reliance on a single global representation limits the impact of quantization errors, which are mainly concentrated in attention, GELU, and embedding layers. When moving from classification to segmentation frameworks new challenges emerge. In particular, decoder components such as masked attention, multi-scale feature fusion, and upsampling layers not only increase computational complexity but also heighten sensitivity to precision loss, making quantization significantly more delicate.

\subsection{Advances in ViT Quantization}
\label{sec:SOTA_Quant}

Recent advances in ViT quantization cover both PTQ \cite{lin2022fqvit, yuan2024ptq4vit, 10.1145/3503161.3547826, li2023repqvitscalereparameterizationposttraining,wu2024adalog, wu2025aphqvit, wu2025fimaq, HE2025103530, jiang2024adfqvitactivationdistributionfriendlyposttrainingquantization, 10839431} and QAT \cite{10.5555/3600270.3602766, li2023vit, 82d09b1bc7e1495889e283a947338ab4, 10.5555/3666122.3666518, ranjan2025mixqvitmixedprecisionvisiontransformer}, with a strong emphasis on achieving integer execution beyond mixed-precision fallbacks. These efforts also focus on stabilizing the most fragile Transformer operators, namely attention (Softmax), LayerNorm, and GELU, whose sensitivity to low-precision scaling often constrains performance in sub-8-bit settings. Recent studies have examined quantization strategies for classification-oriented ViTs, demonstrating both their promise and remaining challenges \cite{du2024model}.

The nonlinear nature of the GELU activation motivated the development of quantization-friendly approximations. For instance, ShiftGELU \cite{li2023vit} approximates GELU with shift-based integer operations, PackQViT \cite{10.5555/3666122.3666518} implements Integer-GELU for hardware efficiency, and APHQ-ViT \cite{wu2025aphqvit} introduces an MLP reconstruction strategy that replaces GELU with ReLU during calibration, enabling stable post-training quantization at ultra-low bitwidths. To mitigate the challenges of quantizing Softmax in the attention layers, FQ-ViT \cite{lin2022fqvit} introduces a Log-Int-Softmax approximation that preserves score monotonicity, while APQ-ViT \cite{10.1145/3503161.3547826} and AdaLog \cite{wu2024adalog} adapt scaling to maintain the power-law structure of attention maps. I-ViT \cite{li2023vit} proposes Shiftmax, a fully integer Softmax based on power-of-two shifts, and PackQViT \cite{10.5555/3666122.3666518} develops Int-2$^{n}$-Softmax, a base-2 integer formulation optimized for SIMD execution. Finally, Q-ViT \cite{10.5555/3600270.3602766} addresses quantization-induced distortions by distilling attention maps from a full-precision teacher. In parallel, PTQ4ViT \cite{yuan2024ptq4vit} specifically targets post-Softmax and post-GELU activations. It introduces Twin Uniform Quantization (TUQ), which applies distinct scaling factors to different value ranges, together with a Hessian-guided metric for more accurate parameter selection, enabling near-lossless accuracy at 8-bit and significant gains at lower bitwidths.

LayerNorm has emerged as one of the most challenging operators to quantize due to its sensitivity to variance estimation. To address this, FQ-ViT introduces a Power-of-Two Factorization to stabilize scaling factors, I-ViT proposes an integer-only reformulation (I-LayerNorm), RepQ-ViT \cite{li2023repqvitscalereparameterizationposttraining} reparameterizes post-normalization scaling to mitigate saturation, and PackQViT  integrates a hardware-oriented Integer-LayerNorm optimized for packed 4-bit arithmetic. FIMA-Q \cite{wu2025fimaq} tackles quantization instability at a global level by improving block-wise reconstruction through a Fisher Information Matrix-based loss. However, subsequent studies have emphasized that it is often not the normalization operation itself, but rather the post-LayerNorm activations, that pose the greatest challenge due to long-tailed distributions and systematic outliers. This insight has motivated outlier-resilient strategies such as ORQ-ViT \cite{HE2025103530} and ADFQ-ViT \cite{wu2025aphqvit}, which specifically target post-LayerNorm activations to enhance robustness under ultra-low-bit quantization. For example, ADFQ-ViT introduces a Per-Patch Outlier-aware Quantizer to selectively retain full-precision outliers, combined with a Shift-Log2 Quantizer to handle asymmetric activations. In line with this direction, PackQViT further revisits the distribution of activations within the ViT dataflow, highlighting the role of long-tailed distributions and systematic channel-wise outliers. To mitigate these effects, it adopts log2-based quantization or clipping for long-tailed activations, and introduces outlier-aware training for residual link quantization, enabling consistent handling of outliers without runtime adaptivity. 
Altogether, advances in ViT quantization show a clear trend toward full integer execution, driven by operator-specific approximations and outlier-resilient strategies, though stabilizing ultra-low-bit activations remains challenging. Thus far, most evaluations have focused on image classification, leaving open questions about their effectiveness and transferability to dense prediction tasks such as semantic segmentation.

%% file: 3_methodology.tex
\section{Methodology}\label{ref_methodology}

Our proposed model, I-Segmenter, is an integer-only adaptation of Segmenter, inspired by the quantization methodology of I-ViT. It quantizes all layers and parameters, representing both linear and non-linear operations exclusively thought bit-shifting and integer arithmetic. We select Segmenter as our architectural foundation due to its structural simplicity and competitive segmentation performance. This model is built on a fully transformed-based pipeline, composed of a standard ViT encoder and a lightweight attention-based decoder that utilizes class-specific mask embeddings. Segmenter employs the same Transformer block in both the encoder and decoder. This architectural homogeneity yields a concise computational graph and improves compatibility with inference engines limited to a small set of operators. In contrast to ViT-based segmentation models with complex modules, Segmenter’s decoder adopts a simple design that avoids costly arithmetic and interpolation, ensuring better support for integer-only and low-precision hardware. Taken together, these properties make Segmenter an ideal testbed for exploring the feasibility of integer-only ViT segmentation and evaluating its compatibility with real-world deployment constraints. 

\subsection{I-Segmenter Overview}

Segmenter's encoder follows the standard ViT architecture, processing non-overlapping image patches and capturing long-range dependencies via self-attention. After $L$ Transformer layers, the input image is represented as a contextualized sequence of patch embeddings $z_L \in \mathbb{R}^{N \times D}$, where $N$ is the number of image patches and $D$ is the embedding dimensionality. The decoder, implemented as a Mask Transformer inspired by DETR \cite{10.1007/978-3-030-58452-8_13}, takes $z_L$ together with a set of learnable class embeddings as input, and produces dense segmentation masks through attention-based refinement. In the proposed I-Segmenter, the entire Segmenter architecture, thus both the encoder and the decoder, is replaced with quantized layers to achieve true integer-only inference. This comprehensive quantization applies to all parameters, including weights, biases, and embeddings (class, and positional) using symmetric uniform quantization, defined as follows:

\begin{equation}
	I = \left\lfloor\frac{\text{clip}(F,-m,m)}{S}\right\rceil, \; \text{where} \,\, S = \frac{2m}{2^k-1}
  \label{eq:quantization}
\end{equation}

where $F$ is the full-precision (FP32) tensor, $m$ is the clipping threshold estimated via an exponential moving average (EMA) of min/max values, $k$ is the quantization bit-width, and $I$ is the resulting $k$-bit integer tensor. Here the scaling factor $S$ maps the real interval $[-m, m] \subset \mathbb{R}$ onto the integer set $\{-2^{k-1}, \ldots, 2^{k-1}-1\} \subset \mathbb{Z}$, ensuring symmetric representation around zero. The operator $\lfloor \cdot \rceil$ denotes nearest integer rounding. The clipping threshold $m$ is updated online across training batches using the EMA update rule:

\begin{align}
  m_{\text{min}, i} &= \alpha \cdot \min(F) + (1-\alpha) \cdot m_{\text{min}, i-1}, \\
  m_{\text{max}, i} &= \alpha \cdot \max(F) + (1-\alpha) \cdot m_{\text{max}, i-1}, \\
  m &= \max(-m_{\text{min}, i},\, m_{\text{max}, i}),
  \label{eq:minmax}
\end{align}
where $\alpha$ is the momentum factor. To enable integer-only inference, $S$ is approximated using dyadic arithmetic, \textit{i.e.}, a ratio of an integer numerator and a power of two denominator:
\begin{equation}
  I \cdot S \;\approx\; I \cdot \frac{b}{2^c} \;=\; (I \cdot b) \gg c,
  \label{eq:dyadic_scaling}
\end{equation}
where $b$ and $c$ are integers obtained through fixed point arithmetic such that $S \approx b/2^c$, and $\gg$ denotes right bit-shifting. This approximation allows scaling operations to be implemented as an integer multiplication followed by a bit-shift, thereby preserving true integer-only execution.

We build on the I-ViT quantization scheme, extending this approach from image classification to the segmentation domain. We further improve I-ViT’s integer-only GELU approximation by introducing an additional parameter $\lambda$ to relax the lower bound of the input distribution. This enhancement enables highly efficient PTQ on top of QAT. To fully support integer-only inference, we make several decoder-side modifications, including the removal of L2 normalization and replacement of the upsampling operator. As the vast majority of I-Segmenter's layers propagate INT8 tensors, our model offers a lightweight alternative for semantic segmentation, well suited to resource-constrained environments. 

\subsection{ViT Encoder Quantization Scheme}
\label{sec:IViT}

The encoder's full computational graph comprises six key layer types that require quantization. They are categorized into two groups: linear operators (Linear, Conv, MatMul) and non-linear ones (LayerNorm, GELU, Softmax). Following the quantization scheme proposed by I-ViT, linear layers are quantized using a dyadic arithmetic pipeline that combines integer scaling with bit-shift operations to approximate full-precision behavior. Non-linear layers are implemented with integer-only approximations, specifically replacing standard activations with Shiftmax, ShiftGELU, and I-LayerNorm. In line with industry standards \cite{Jacob_2018_CVPR}, we apply symmetric uniform quantization with distinct precision for weights (INT8) and biases (INT32).
The product of two INT8 values (input activations and weights) is accumulated into an INT32 register. Consequently, biases are added to the result before the tensor is requantized back to INT8. To preserve precision, residual connections are computed by adding two INT16 tensors with an INT32 accumulator, and the outcome is then requantized to INT16. Maintaining residuals in INT16 within Transformer block preserves precision and ensures consistent scaling at each skip connection. Furthermore, the most precision-sensitive operation, Shiftmax, is carried out in INT16 precision, to preserve accuracy in the attention mechanism's $\text{Shiftmax}(QK^T) \cdot V$ computation. Figure~\ref{fig:I-Segmenter_TransformerBlock} shows a zoomed-in view of a Transformer block used in the I-Segmenter encoder and reused in the decoder. It illustrates the sequence of quantized operations from left to right, following the data flow through the attention and feed-forward sub-blocks, and highlights how numerical precisions are propagated and adjusted via requantization steps. Operators inherited from the I-ViT quantization scheme and those introduced in our method are distinguished by color.

While the quantized Shiftmax and I-LayerNorm layers retain accuracy, we find that the discrepancy between FP32 GELU and I-ViT’s integer-only approximation (ShiftGELU) is too large to enable effective PTQ or stable QAT. I-ViT builds on previous work on GELU approximations \cite{hendrycks2016gaussian} to derive an integer-only variant, formally defined as:

\begin{equation}
\begin{aligned}
  \text{GELU}(x) &\approx S_x \cdot I_x \cdot \, \sigma(S_x \cdot 1.702\, I_x),
\end{aligned}
\label{eq:GELU}
\end{equation}
\noindent
where $I_x$ the integer vector, $S_x$ is the corresponding scaling factor, and $\sigma$ is the sigmoid function. To avoid floating-point multiplications, the constant $1.702$ (value derived in~\cite{hendrycks2016gaussian}) is approximated by its binary expansion $(1.1011)_2 \approx 1.6875$, which can be implemented using shift-and-add operations:
\begin{equation}
\begin{aligned}
  I_p &= 1.702\, I_x \\
  &\approx I_x + (I_x \gg 1) + (I_x \gg 3) + (I_x \gg 4),
\end{aligned}
\end{equation}
\noindent
The sigmoid function is further expressed as
\begin{equation}
\begin{aligned}
  \sigma(S_x \cdot I_p) &= \frac{1}{1 + e^{-S_x \cdot I_p}} \\
  &= \frac{e^{S_x \cdot (I_p - I_{\max})}}{e^{S_x \cdot (I_p - I_{\max})} + e^{-S_x \cdot I_{\max}}},
\end{aligned}
\label{eq:sigmoid}
\end{equation}
with $I_{\max} = \max\{I_{p1}, I_{p2}, \ldots, I_{pd}\}$. I-ViT uses $I_{max}$ to smooth the data distribution and prevent overflow. To simplify notation, we will refer to $I_p - I_{max}$ as $I_{\Delta}$. Exponentials are approximated through base conversion:
\begin{equation}
\begin{aligned}
  e^{S_x I_\Delta} &= 2^{S_{x}(I_\Delta \text{log}_{2} e)} \\
  &\approx 2^{S_x I_e}, \\
  I_e &= I_\Delta + (I_\Delta \gg 1) - (I_\Delta \gg 4).
\end{aligned}
\label{eq:change_base}
\end{equation}
\noindent
Bit-shifting operations are used to provide an efficient integer approximation of $\log_2 e \approx 1.4427$. In practice, I-ViT clamps $I_e$ to prevent overflow:
\begin{equation}
  I_e = \texttt{clamp\_min}\!\left(I_e,\, k_{\text{inter}} \cdot (-I_0)\right),
\label{eq:lambda_shift_gelu}
\end{equation}
\noindent
where $I_0 = \lfloor 1/S_x \rceil$, and $\texttt{clamp\_min}(x, a)$ denotes an element-wise operation that enforces $x \geq a$ by replacing values below $a$ with $a$. The variable $k_{\text{inter}}$ denotes the bit-precision used for intermediate computations.

We observe that this clamping operation is overly restrictive, as it suppresses informative negative values and introduces significant downstream errors. To address this, we introduce a tunable scalar parameter $\lambda$ that relaxes the lower bound:
\begin{equation}
  I_e = \texttt{clamp\_min}\!\left(I_e,\, \lambda \cdot k_{\text{inner}} \cdot (-I_0)\right),
\end{equation}
\noindent
yielding a more robust variant, which we denote as $\lambda\text{-ShiftGELU}$ (see~\ref{app1} for the pseudo-code). The choice of $\lambda$ is closely tied to the internal precision $k_{\text{inter}}$, as it governs the dynamic range of intermediate computations. To ensure the power term $S_x I_e$ is an integer, I-ViT decomposes it into an integer part $q$ and a decimal part $r$ as follows:
\begin{equation}
\begin{aligned}
  2^{S_x I_e} &= 2^{(k_{inter} - q) + S_x(-r)} \\
  &= 2^{S_x(-r)} \ll (k_{inter} - q) \\
  &\approx (S_x [((-r) \gg 1) + I_0]) \ll (k_{inter} - q),
\end{aligned}
\label{eq:decimal_and_integer}
\end{equation}
\noindent
where $q$ and $r$ are integers, and $\ll$ denotes the left bit-shift operator. To prevent excessively small values, the right shift by $q$ is moderated using $k_{\text{inter}}$. Setting $k_{\text{inter}}$ too low results in a loss of precision for intermediate computations and therefore significant accuracy degradation. On the other hand, setting $k_{\text{inter}}$ too high is inefficient; maximal precision could have been obtained with a lower value and therefore fewer bit-shifting operations. Through manual hyperparameter tuning, we experimentally find $\lambda = 6$ and $k_{\text{inter}} = 23$ strikes this balance point. Finally, division inside the sigmoid is replaced by an integer-only operator:
\begin{equation}
\begin{aligned}
  I_{div} &= \text{IntDiv}(I_{\exp}, I_{\exp} + I'_{\exp}, k_{\text{out}}) \\
  &= \left(\Big\lfloor\frac{2^{31}}{I_{\exp} + I'_{\exp}}\Big\rfloor \cdot I_{\exp}\right) \gg \big(31-(k_{\text{out}}-1)\big),
\end{aligned}
\label{eq:IntDiv}
\end{equation}
\noindent
where $k_{\text{out}}$ sets the output scale, $I_{exp} \approx e^{S_x \cdot (I_p - I_{\max})}$, and $I'_{exp} \approx e^{-S_x I_{\max}}$. The corresponding scaling factor is $S_{div} = 2^{-(k_{out} - 1)}$. In addition to the new $\lambda$-ShiftGELU operator, we introduce several modifications to the encoder. Specifically, we remove the Dropout and DropPath regularization layers, as these stochastic operations are unnecessary for our quantization approach and introduce variability incompatible with integer-only inference. Unlike I-ViT, we also quantize the class and positional embeddings, ensuring integer-only computation. Crucially, all weights, biases, embeddings, and scaling factors can be stored in integer form (through dyadic representation), so that the model is integer-only not just in execution but also in storage.

\subsection{I-Segmenter's Decoder: I-MaskTransformer}
\label{sec:IMaskTransformer}

\begin{figure}[!t]
\centering
\includegraphics[width=\columnwidth]{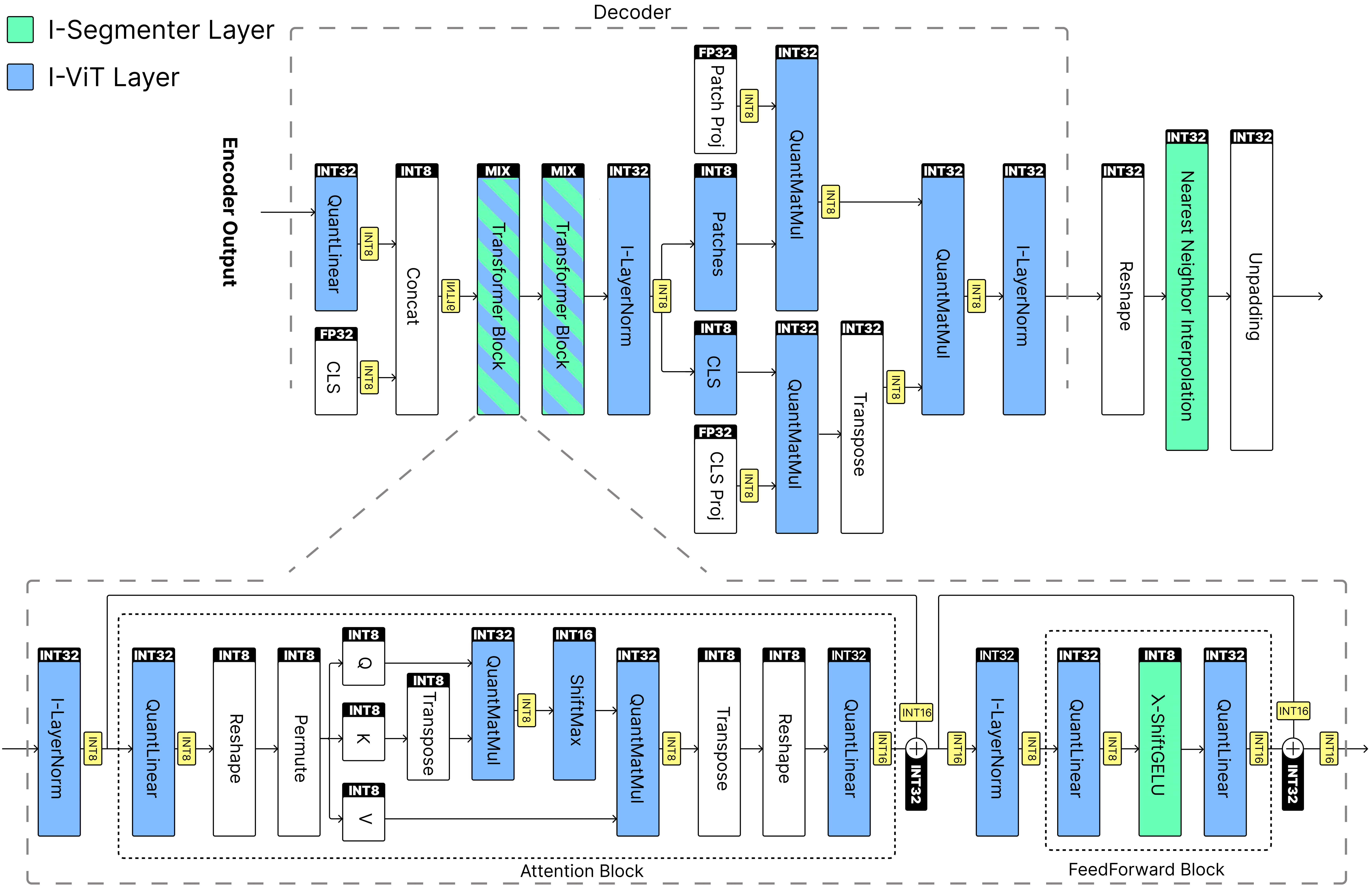}
\caption{Overview of the proposed I-MaskTransformer, featuring an integer-only decoder head integrated into the I-Segmenter architecture. The encoder and decoder share an identical Transformer block. Green and blue highlight, respectively, the operators introduced by I-Segmenter and by the I-ViT quantization scheme. Requantization operations are represented by yellow stumps.}
\label{fig:I-Segmenter_TransformerBlock}
\end{figure}

The Mask Transformer is responsible for generating a set of $K$ class masks for dense prediction using the output of the encoder. Unlike standard classification settings where a single class token is used, the Mask Transformer replaces the encoder's class token with a set of learnable class embeddings, defined as $\text{cls} = [\text{cls}_1, \dots, \text{cls}_K] \in \mathbb{R}^{K \times D}$. These embeddings are processed alongside the contextualized patch representations $z_L$ obtained from the encoder. Following the same architecture as the encoder, the Mask Transformer applies two additional Transformer blocks to jointly process the class embeddings and the patch tokens. This produces class-aware features: updated patch tokens $z_M \in \mathbb{R}^{N \times D}$ and refined class embeddings $c \in \mathbb{R}^{K \times D}$. To enrich the representations before mask generation, the patch tokens and class embeddings are projected through learnable matrices $P_M \in \mathbb{R}^{N \times N}$ and $P_c \in \mathbb{R}^{K \times K}$, respectively. This yields $z'_M \in \mathbb{R}^{N \times D}$ and $c' \in \mathbb{R}^{K \times D}$, which are smoothened out through L2 normalization. Finally, class-specific masks are obtained via the inner product between the projected patches and transposed class embeddings:
\begin{equation}
  \text{Masks}(z_{M}', c') = z_{M}'c'^T.
\end{equation}

\noindent Once class-specific masks are produced at patch resolution, Segmenter refines them through reshaping, bilinear interpolation, and unpadding so that the predicted masks align precisely with the pixels of the input image.

I-Segmenter's decoder adopts the same mapping from full-precision to quantized operators as in the encoder, enabling the first fully integer-only Transformer decoder for segmentation, referred to as I-MaskTransformer. The upper part of Figure~\ref{fig:I-Segmenter_TransformerBlock} provides an overview of the architecture and quantization pipeline, illustrating how floating-point operations are replaced with integer-compatible counterparts. We introduce several adjustments to make the Segmenter pipeline compatible with integer-only inference. In particular, we omit L2 normalization of $z'_M$ and $c'$ prior to computing the inner product. Given that we requantize $z'_M$ and $c'$ after matrix multiplication. L2 normalization introduces more complex quantization procedures that require non-trivial approximations, which are undesirable in an integer-only settings. Therefore, instead of L2 normalization, we use symmetric uniform quantization to control the ranges of both matrices and smooth their distributions. Finally, we replace the upsampling operator, which relies on floating-point interpolation, with nearest neighbor interpolation. Nearest neighbor interpolation uses simple index rounding to select the closest pixel, and is possibly the simplest upsampling method compatible with integer arithmetic which is already implemented in most inference engines.

%% file: 4_experimental_setup.tex
\section{Experimental Setup}

We evaluate I-Segmenter along five dimensions: (1) segmentation accuracy, reported as mean Intersection over Union (mIoU, \%); (2) inference latency, measured as the end-to-end runtime for a single sample (milliseconds); (3) memory footprint, quantified as the total bit-level read/write volume; (4) model size, reported as the checkpoint size (MB); and (5) training and calibration time, recorded in seconds or hours. Comparisons are made against the baseline FP32 Segmenter model across large-scale benchmarks: ADE20K \cite{zhou2017scene} and Cityscapes \cite{Cordts2016Cityscapes}. As the encoder is based on the plain ViT architecture, we consider the same four classic architecture sizes: Tiny, Small, Base, and Large. Segmenter checkpoints were either retrieved from the ALGM repository \cite{norouzi2024algm}, as the original releases are no longer available, or generated by training the model from scratch. All experiments are conducted on systems equipped with NVIDIA A100 GPUs (40GB). While training may leverage multiple GPUs for efficiency, all reported training times and memory metrics correspond to a single GPU. Inference and latency evaluations are performed on a single A100 GPU across all frameworks to ensure fair comparison.

\subsection{Training and Calibration Strategy}

To enable efficient inference with minimal accuracy loss, we explore both PTQ and QAT as strategies for compressing I-Segmenter. While QAT can recover accurate activation and weight distributions through FP32 gradient updates, PTQ lacks this adaptability, as it relies solely on calibrating scaling factors without modifying model parameters.

In QAT, we fine-tune the pre-trained Segmenter checkpoint using a standard pixel-wise cross-entropy loss, a polynomial learning rate scheduler, and an SGD optimizer. We use the LR range test \cite{smith2017cyclical} to identify optimal learning rates. We train ADE20k models for 64 epochs and Cityscapes models for 216 epochs, with the first 10 epochs used for warmup in both cases. After loading checkpoints, we employ JIT compilation to run I-Segmenter PyTorch code with optimized kernels and use automatic mixed precision (AMP) to promote BF16 precision instead of FP32, further reducing resource consumption. Since computing scaling factors with min/max EMA and fixed-point approximation is expensive, we set the momentum factor $\alpha$ very low, to 0.01, and freeze EMA after the first 10 warmup epochs. 

In PTQ, the model weights remain frozen while activation scaling factors are calibrated dynamically. Consequently, PTQ is inherently more limited in compensating for quantization-induced distortions. Following I-ViT, we set $\alpha = 0.05$, giving more weight to recent calibration samples while gradually reducing the influence of earlier ones, which are effectively forgotten after approximately 100 iterations. We explore one-shot calibration and compare it to PTQ using a calibration set of size $S=500$ (batch size set to 8), which in our experiments is sufficient for the EMA to converge. This analysis allows us to examine the trade-off between precision, as provided by QAT, and resource efficiency, as enabled by PTQ, while highlighting the performance of I-Segmenter under extreme conditions such as one-shot calibration.

For data preprocessing, we adopt the standard pipeline from the MMSegmentation library \cite{mmseg2020}, including random left-right flipping, random resizing with a scale factor between 0.5 and 2.0, and mean subtraction. Small images are padded to a fixed size of $768 \times 768$ for Cityscapes and $512 \times 512$ for ADE20K, while larger images are randomly cropped to the same target sizes.

\subsection{Execution Backends for Quantized Models Workflow}

\begin{figure*}[!ht]
\centering
\includegraphics[width=1.0\linewidth]{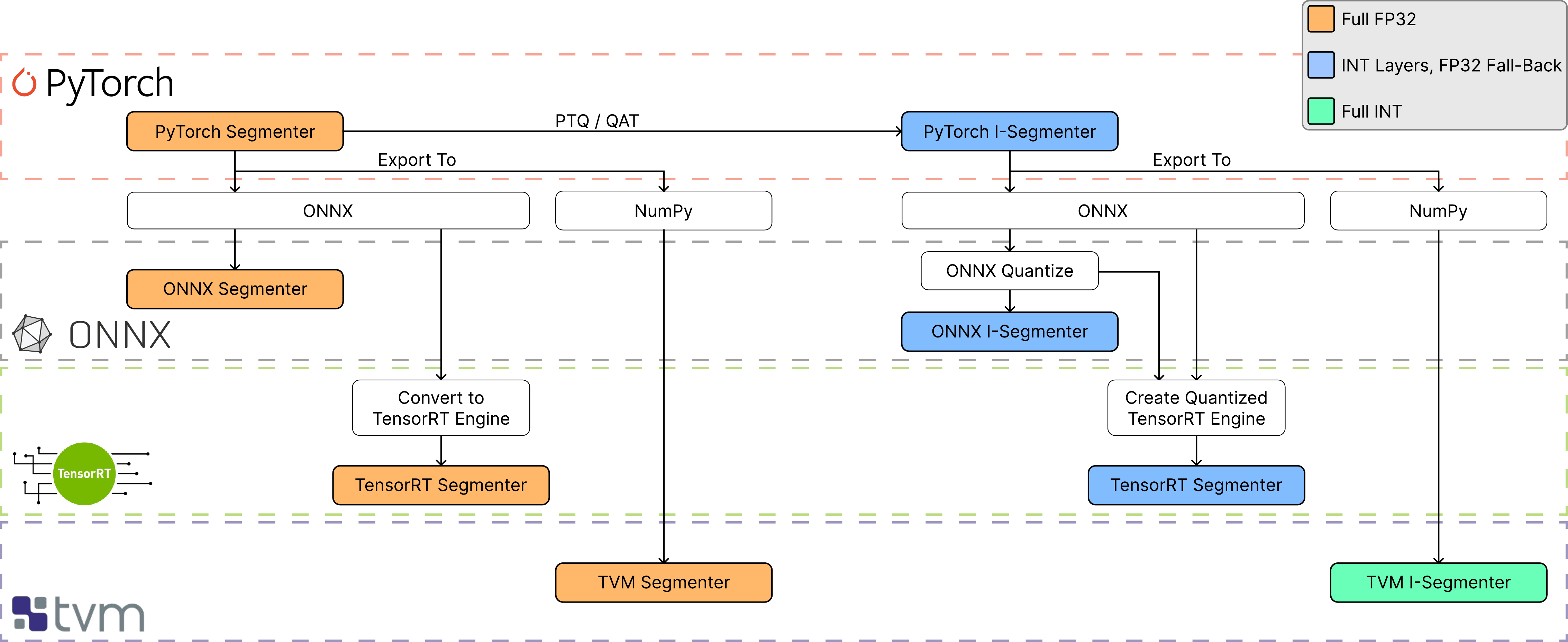}
\caption{ Inference pipeline of I-Segmenter, from the PyTorch model to ONNX, TensorRT, and TVM backends, with FP32 and INT support. All models are derived from PyTorch. To obtain quantized I-Segmenter in TensorRT, we consider the traditional path directly from FP32 ONNX, and from a pre-quantized ONNX export. The only backend to achieve true INT only inference down to kernel computations is TVM.}
\label{fig:Inf_beckends}
\end{figure*}

We evaluate Segmenter and I-Segmenter across three backends: ONNX Runtime and TensorRT, which target high-performance production inference, and TVM, which emphasizes flexibility and kernel-level customization. For TensorRT export, we consider two pathways, starting from an FP32 I-Segmenter checkpoint or from a pre-quantized ONNX Runtime checkpoint (denoted as ONNX-TensorRT). Figure~\ref{fig:Inf_beckends} illustrates the complete pipeline used to generate our inference models. Although ONNX Runtime and TensorRT support quantization out-of-the-box, they fail to enforce integer-only execution throughout the entire model, frequently falling back to FP32 operations. This limitation arises from incomplete support for integer-only operators and the black-box nature of their quantization strategies, which prevent fine-grained control over kernel precision. To overcome this limitation, we leverage TVM’s manual kernel control to enforce strict integer-only operators across the entire model. Translating our PyTorch implementation into TVM required substantial engineering efforts and resulted in a minor decrease in accuracy.

%% file: 5_ablation.tex
\section{Ablation studies}

\subsection{%
  \texorpdfstring{Integer-only $\lambda$-ShiftGELU}{Integer-only lambda-ShiftGELU}
}
\label{sec:Lambda_GELU_ablation}

We study the effect of scaling the lower bound for distribution clamping in the bit-shifting approximation of the exponential function ShiftGELU using the parameter $\lambda$. To quantify the discrepancy between integer-only GELU approximations and the FP32 reference, we employ the global root mean squared error ($\text{RMSE}_{\text{G}}$), averaged across all Transformer layers and over the entire dataset. We formally define it as follows:

\begin{equation}
\text{RMSE}_{\text{G}} = 
\sqrt{ \frac{1}{T \cdot N \cdot D} 
\sum_{l=1}^{T} \sum_{i=1}^{N} \sum_{j=1}^{D} 
\left( f^{(l)}(x_i)_j - \hat{f}^{(l)}(x_i)_j \right)^2 },
\end{equation}

\noindent
where $T = L_{\text{enc}} + L_{\text{dec}}$ is the total number of Transformer blocks in the model, $N$ is the number of samples in the dataset, and $D$ is the embedding dimensionality.

\begin{table}[t]
\centering
\footnotesize
\begin{tabular}{l l c c c c}
\toprule
\textbf{Dataset} & \textbf{GELU Type} &
\textbf{Tiny} & \textbf{Small} & \textbf{Base} & \textbf{Large} \\
\midrule
\multirow{2}{*}{\textbf{ADE20k}}
 & ShiftGELU            & 4.52 & 414.24 & 373.67 & 293.94 \\
 & $\lambda$-ShiftGELU  & 4.35 & 14.44  & 8.83   & 3.75   \\
\midrule
\multirow{2}{*}{\textbf{Cityscapes}}
 & ShiftGELU            & 4.19 & 401.91 & 308.68 & 300.78 \\
 & $\lambda$-ShiftGELU  & 3.54 & 14.30  & 7.00   & 9.28   \\
\bottomrule
\end{tabular}
\caption{Comparison of GELU approximation fidelity between $\lambda$-ShiftGELU and ShiftGELU, measured by RMSE across different backbones in the one-shot PTQ setting.}
\label{tab:Quantitative_impact_lambdaShiftGelu}
\end{table}

\begin{figure}[H]
\centering
\begin{minipage}{0.48\textwidth}
    \centering
    \includegraphics[width=\linewidth]{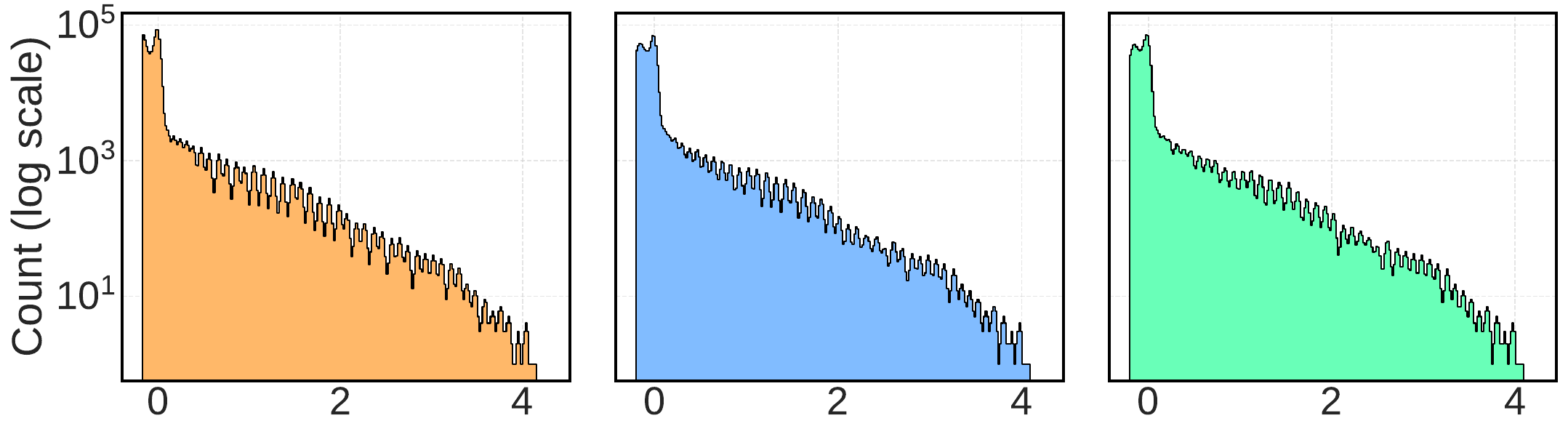}
\end{minipage}
\hfill
\begin{minipage}{0.48\textwidth}
    \centering
    \includegraphics[width=\linewidth]{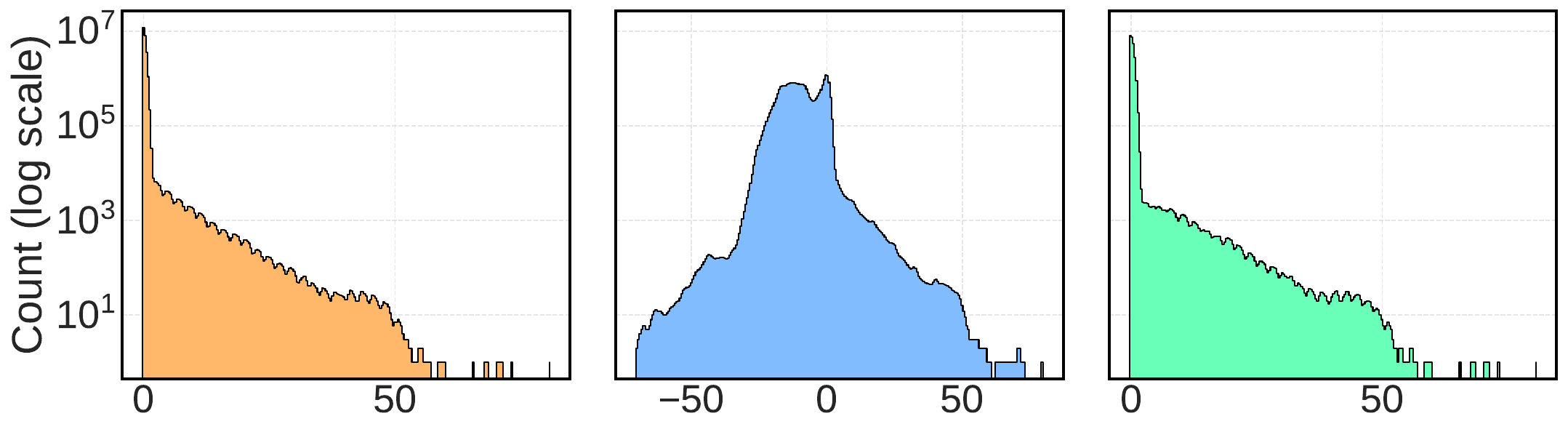}
\end{minipage}
\caption{GELU distribution comparison in the 10th encoder transformer block of the ViT backbone for a ViT-Tiny model (left) and a ViT-Base model (right). The encoder consists of 12 blocks, which are counted sequentially. From left to right: FP32 GELU (baseline), INT8 ShiftGELU from I-ViT, and our proposed INT8 $\lambda$-ShiftGELU.}
\label{fig:Impact_lambda}
\end{figure}


QAT provides an additional 1--3 mIoU points of improvements over PTQ.As shown in Table~\ref{tab:Quantitative_impact_lambdaShiftGelu}, in the one-shot PTQ setting, $\lambda$-ShiftGELU consistently provides a more faithful approximation of the activation function, reducing the global RMSE by up to 98.7\% compared to the baseline ShiftGELU. An exception is observed in I-Segmenter Tiny, where both GELU variants approximate the distribution equally well. A possible reason is that the smaller embedding dimension reduces activation variance and the strength of the GELU non-linearity, making ShiftGELU sufficient, while larger models amplify these effects and therefore benefit more from $\lambda$-ShiftGELU. A visual comparison in Figure~\ref{fig:Impact_lambda} illustrates the output distributions of GELU, ShiftGELU, and $\lambda$-ShiftGELU for the same input. For larger backbones such as ViT-Base, ShiftGELU deviates substantially from the baseline GELU, whereas $\lambda$-ShiftGELU closely aligns with its distribution and exhibits nearly identical trends.

\subsection{Effect of Interpolation Choice and Normalization in Mask Refinement}

We investigate the impact of two key decoder-side modifications: replacing bilinear upsampling with nearest-neighbor interpolation, and removing the L2-normalization layer. To assess their effect, we measure the mIoU of I-Segmenter in the one-shot PTQ setting under different configurations. Results are reported in Table~\ref{tab:Interpolation_and_L2Norm}. As expected, the highest mIoU scores are obtained when both bilinear interpolation and L2 normalization are preserved. In fact, nearest-neighbor interpolation introduces discretization artifacts, resulting in pixelation and jagged boundaries. These effects are particularly pronounced at object boundaries and in fine-grained regions, where both PTQ and QAT exhibit staircase-like edges and fragmented predictions (see Figure~\ref{fig:appendix_predictions} in \ref{sec:app2}). Removing L2 normalization introduces a consistent, though relatively minor, degradation of approximately 0.9 mIoU points on average, suggesting that normalization through requantization slightly perturbs the output distribution. However, introducing fixed-point arithmetic to replace L2 normalization comes at significant modeling and implementation complexity, therefore we opted for removing the operation completely and ensuring strict integer-only inference. By contrast, the interpolation choice has a more pronounced effect. Replacing bilinear with nearest-neighbor interpolation leads to a substantial accuracy drop, averaging 1.7 mIoU points. While computationally simpler and integer-friendly, nearest-neighbor interpolation sacrifices the ability to capture smooth and nuanced boundaries, effectively mapping all pixels within an upsampled region to the same class. The observed gap, averaging 3.4 mIoU points, reflects the current performance cost of enforcing strict integer portability. Improvements to the normalization and interpolation steps are left to future work.

\begin{table}[t]
\centering
\footnotesize
\begin{tabular}{l l c c c c c}
\toprule
\textbf{Dataset} & \textbf{Interp.} & \textbf{L2 Norm} &
\textbf{Tiny} & \textbf{Small} & \textbf{Base} & \textbf{Large} \\
\midrule
\multirow{4}{*}{\textbf{ADE20k}}
 & Nearest  & No  & 32.60 & 40.87 & 45.43 & 47.46 \\
 & Nearest  & Yes & 35.89 & 42.52 & 45.82 & 47.88 \\
 & Bilinear & No  & 32.56 & 41.55 & 46.38 & 47.75 \\
 & Bilinear & Yes & 36.59 & 43.70 & 47.47 & 49.85 \\
\midrule
\multirow{4}{*}{\textbf{Cityscapes}}
 & Nearest  & No  & 68.31 & 71.98 & 71.85 & 72.90 \\
 & Nearest  & Yes & 68.74 & 72.44 & 72.28 & 73.15 \\
 & Bilinear & No  & 70.46 & 75.09 & 75.24 & 75.90 \\
 & Bilinear & Yes & 72.03 & 76.07 & 76.17 & 76.91 \\
\bottomrule
\end{tabular}
\caption{Impact of interpolation method and L2-normalization on performance.}
\label{tab:Interpolation_and_L2Norm}
\end{table}

%% file: 6_results.tex
\section{Results}
\label{sec:Results}

\subsection{Accuracy Evaluation}
\label{sec:Accuracy}

\begin{table}[t]
\centering
\footnotesize
\begin{tabular}{l c l c c c c}
\toprule
\textbf{Model} & \textbf{S} & \textbf{GELU Type} &
\textbf{Tiny} & \textbf{Small} & \textbf{Base} & \textbf{Large} \\
\midrule
\textbf{Baseline} \\
Segmenter & -- & GELU & 38.23 & 45.89 & 49.28 & 52.29 \\
\midrule
\textbf{PTQ} \\
I-Segmenter & 1   & ShiftGELU           & 32.06 & 0.92  & 1.48  & 0.15 \\
I-Segmenter & 500 & ShiftGELU           & 32.39 & 3.70  & 2.28  & 0.15 \\
I-Segmenter & 1   & $\lambda$-ShiftGELU & 32.60 & 40.87 & 45.43 & 47.46 \\
I-Segmenter & 500 & $\lambda$-ShiftGELU & 32.66 & 40.15 & 43.67 & 47.82 \\
\midrule
\textbf{QAT} \\
I-Segmenter & -- & $\lambda$-ShiftGELU & 37.62 & 43.44 & 46.19 & 48.74 \\
\bottomrule
\end{tabular}
\caption{Accuracy analysis (mIoU) of I-Segmenter on ADE20K with PTQ and QAT.}
\label{tab:mIoU_ade20k}
\end{table}


\begin{table}[t]
\centering
\footnotesize
\begin{tabular}{l c l c c c c}
\toprule
\textbf{Model} & \textbf{S} & \textbf{GELU Type} &
\textbf{Tiny} & \textbf{Small} & \textbf{Base} & \textbf{Large} \\
\midrule
\textbf{Baseline} \\
Segmenter & -- & GELU & 73.48 & 77.09 & 77.64 & 79.06 \\
\midrule
\textbf{PTQ} \\
I-Segmenter & 1   & ShiftGELU           & 67.48 & 22.84 & 10.30 & 3.92 \\
I-Segmenter & 500 & ShiftGELU           & 67.82 & 26.69 & 11.81 & 5.70 \\
I-Segmenter & 1   & $\lambda$-ShiftGELU & 68.31 & 71.98 & 71.85 & 72.90 \\
I-Segmenter & 500 & $\lambda$-ShiftGELU & 67.90 & 72.05 & 70.54 & 73.35 \\
\midrule
\textbf{QAT} \\
I-Segmenter & -- & $\lambda$-ShiftGELU & 70.23 & 73.28 & 73.23 & 74.09 \\
\bottomrule
\end{tabular}
\caption{Accuracy analysis (mIoU) of I-Segmenter on Cityscapes with PTQ and QAT.}
\label{tab:mIoU_cityscapes}
\end{table}

\begin{table}[t]
\centering
\footnotesize
\begin{tabular}{l c c c c c}
\toprule
\textbf{Model} & \textbf{S} &
\textbf{Tiny} & \textbf{Small} & \textbf{Base} & \textbf{Large} \\
\midrule
\textbf{Training / Calibration Time} \\
I-Segmenter (PTQ) & 1   & 1\,s  & 1\,s  & 1\,s  & 1\,s  \\
I-Segmenter (PTQ) & 500 & 20\,s & 26\,s & 37\,s & 82\,s \\
I-Segmenter (QAT) & --  & 8\,h  & 14\,h & 26\,h & 70\,h \\
\midrule
\textbf{Peak GPU Usage (GB)} \\
I-Segmenter (PTQ) & 1   & 1.76 & 2.02 & 2.86 & 4.86 \\
I-Segmenter (PTQ) & 500 & 5.61 & 6.59 & 8.91 & 12.88 \\
I-Segmenter (QAT) & --  & 38.16 & 36.48 & 35.18 & 33.54 \\
\bottomrule
\end{tabular}
\caption{Time and memory cost of PTQ versus QAT on ADE20K.}
\label{tab:table_qantif_time}
\end{table}

Accuracy of PTQ and QAT I-Segmenter variants is evaluated against the full-precision Segmenter baseline, with all mIoU scores obtained in PyTorch. On the ADE20k dataset, Table~\ref{tab:mIoU_ade20k}, PTQ performs remarkably well, with at most a 5.6 point drop. QAT I-Segmenter narrows this gap further, with an accuracy reduction not exceeding 3.5 mIoU points. A similar trend is observed on Cityscapes, Table~\ref{tab:mIoU_cityscapes}, where PTQ is within 5-7 mIoU points of the baseline, and QAT provides an additional 1-3 mIoU points of improvements over PTQ. Across all datasets, and consistent with our ablation study in Section~\ref{sec:Lambda_GELU_ablation}, $\lambda$-ShiftGELU provides a much closer approximation to the FP32 GELU than I-ViT's ShiftGELU. This improved fidelity directly translates into higher segmentation accuracy: while ShiftGELU collapses under PTQ, $\lambda$-ShiftGELU maintains accuracy close to the full-precision baseline, even with a single calibration sample ($S=1$). The results demonstrate that $\lambda$-ShiftGELU enables effective one-shot calibration: with only a single calibration sample, PTQ achieves mIoU comparable to using $S=500$, drastically reducing calibration cost while preserving segmentation accuracy.
In the~\ref{sec:vis)}, Figure~\ref{fig:appendix_predictions} shows additional qualitative segmentation results on Cityscapes, while Figures~\ref{fig:confusion_matrix} and~\ref{fig:diff_confusion_matrix} compare the confusion matrices of the FP32 Segmenter and the PTQ I-Segmenter. Both models perform similarly on most classes, but PTQ exhibits more errors on small, thin, or visually similar classes, whereas large classes remain largely unaffected (see Figure~\ref{fig:Per-classIoU}).

Table~\ref{tab:table_qantif_time} highlights the drastic difference in computational requirements between PTQ and QAT on ADE20k. 
Calibration with PTQ is extremely lightweight, in the one-shot setting, it completes in about one second regardless of model size, and even with 500 samples, it remains below two minutes for the largest model. In contrast, QAT requires hours of training (from 8h for I-Segmenter Tiny up to 70h for I-Segmenter Large). Memory usage shows a similar trend: PTQ consumes only a few gigabytes of GPU memory (from 1.76~GB for Tiny with $S=1$ up to 12.88~GB for Large with $S=500$), whereas QAT requires more than 30~GB consistently across models.

\subsection{Latency and Memory Evaluation}
\label{sec: Latency_Memory}

\begin{table*}[t]
\centering
\footnotesize
\begin{tabular}{p{1.2cm} c c c c c c c c c}
\toprule
 &  &
\multicolumn{2}{c}{\textbf{Tiny}} &
\multicolumn{2}{c}{\textbf{Small}} &
\multicolumn{2}{c}{\textbf{Base}} &
\multicolumn{2}{c}{\textbf{Large}} \\
\cmidrule(lr){3-4} \cmidrule(lr){5-6} \cmidrule(lr){7-8} \cmidrule(lr){9-10}
\textbf{\shortstack{Inf.\\Backend}} &\textbf{\shortstack{Int.\\Only}} & Size & Lat. & Size & Lat. & Size & Lat. & Size & Lat. \\
\midrule
\multicolumn{10}{c}{\textbf{Segmenter}} \\

PyTorch        & \xmark & 26.41 & 16.15 & 100.84 & 16.71 & 393.99 & 17.38 & 1272.97 & 48.07 \\
ONNX\textsubscript{RT} & \xmark & 25.79 & 20.60 & 99.49  & 24.27 & 391.18 & 24.49 & 1269.26 & 38.93 \\
TRT       & \xmark & 28.04 & 11.41 & 102.79 & 13.21 & 394.72 & 14.98 & 1273.08 & 28.66 \\
TVM            & \xmark & 25.62 & 6.25  & 99.32  & 12.93 & 391.09 & 32.32 & 1268.96 & 105.49 \\
\addlinespace
\multicolumn{10}{c}{\textbf{I-Segmenter}} \\

PyTorch        & \xmark & 26.82 & 111.54 & 101.37 & 111.57 & 394.75 & 118.98 & 1274.63 & 212.39 \\
ONNX\textsubscript{RT} & \xmark & 9.46  & 117.09 & 29.30  & 157.43 & 105.07 & 266.95 & 331.24  & 530.05 \\
TRT       & \xmark & 34.13 & 14.23  & 112.09 & 19.98  & 402.27 & 25.28  & 1290.52 & 52.06 \\
TRT\textsubscript{ONNX} & \xmark & 12.13 & 9.96   & 33.01  & 11.73  & 107.96 & 14.41  & 334.42  & 24.36 \\
TVM            & \cmark & 7.71  & 7.11   & 27.79  & 14.54  & 105.30 & 27.97  & 329.70  & 84.59 \\
\bottomrule
\end{tabular}
\caption{Model size (MB) and end-to-end latency (ms) of Segmenter and I-Segmenter on ADE20K across different inference frameworks. ONNX\textsubscript{RT} denotes ONNX Runtime, TRT denotes TensorRT while TRT\textsubscript{ONNX} refers to the ONNX–TensorRT execution backend.}
\label{table_inferance_Results}
\end{table*}

\begin{table}[t]
\centering
\footnotesize
\begin{tabular}{l l c c c c}
\toprule
\textbf{Metric} & \textbf{Model} &
\textbf{Tiny} & \textbf{Small} & \textbf{Base} & \textbf{Large} \\
\midrule
\textbf{Memory Traffic} \\
 & Segmenter   & 1.01 & 3.08 & 10.36 & 32.00 \\
 & I-Segmenter & 0.43 & 1.13 & 3.31  & 9.74  \\
\midrule
\textbf{mIoU} \\
 & Segmenter   & 37.79 & 44.97 & 48.12 & 50.82 \\
 & I-Segmenter & 37.41 & 41.39 & 45.35 & 47.99 \\
\bottomrule
\end{tabular}
\caption{Memory traffic (total number of bits read and written) and mIoU on ADE20K of I-Segmenter QAT deployed in TVM. Memory traffic was computed as the total number of bits read and written across convolutional, linear, and matmul operations for the given input, output, and accumulation types. The final sum is scaled by $10^{-11}$.}
\label{tab:TVM_inference}
\end{table}

The efficacy of quantization is ultimately reflected in performance on the target hardware. Accordingly, for I-Segmenter we evaluate compression ratio, inference speed, and accuracy across diverse backends. A key distinction among the backends is their adherence to strict integer-only inference. Although ONNX Runtime, TensorRT, and TVM are all derived from the same PyTorch model (see Figure~\ref{fig:Inf_beckends}), their operator support and fusion strategies diverge. TVM uniquely enforces integer-only execution across the entire computational graph, providing native support for quantization operations such as scaling via dyadic arithmetic (integer multiplication and bit-shifting). By contrast, ONNX Runtime and TensorRT frequently revert to FP32 execution when encountering unsupported operator sequences. For example, linear layers may include quantization/dequantization stubs, yet their core computation often falls back to FP32.

As shown in Table~\ref{table_inferance_Results},  I-Segmenter achieves a substantial model compression of $3.2\times$–$3.8\times$ compared to the full-precision baseline across all model variants. The latency benefits of quantization are, however, highly framework-dependent. PyTorch, lacking optimized integer kernels, suffers from significant quantization/dequantization overhead, leading to slower inference. In contrast, TensorRT and TVM effectively exploit quantization to accelerate computation. The ONNX–TensorRT pipeline achieves the fastest runtimes, with I-Segmenter Large reaching a $1.2\times$ speedup over FP32. Moreover, the advantages of quantization scale with model size. In I-Segmenter Base and Large variations, the cost of requantization is amortized by the greater volume of integer arithmetic. In I-Segmenter Tiny, however, this overhead can outweigh the gains, leading to a modest slowdown ($1.1\times$).
Table~\ref{tab:TVM_inference} reports the total number of bits read and written for key layers (Linear, Conv, and MatMul), along with the mIoU of Segmenter and I-Segmenter in TVM. The reduction in bit-level operations by $2.3\times$–$3.3\times$ for I-Segmenter demonstrates its potential for significantly faster and more energy-efficient inference on hardware optimized for integer data types. We evaluate model accuracy after compilation to TVM to isolate degradation introduced by the deployment framework. Compared to PyTorch FP32 inference of the QAT model (Table~\ref{tab:mIoU_ade20k}), we observe a consistent 1-2 point drop in mIoU. This loss is attributed to the manual translation process and differences in backend-specific numerical handling, like rounding, scaling, or accumulation strategies. Despite this, I-Segmenter maintains competitive accuracy, proving that high-performance, pure integer inference is viable without substantial quality loss.

%% file: 7_conclusion.tex
\section{Conclusions}
\label{sec:conslusions}

We introduced I-Segmenter, the first fully integer-only Vision Transformer for semantic segmentation. By systematically replacing all floating-point operators in Segmenter with integer-friendly counterparts, we demonstrated that high segmentation accuracy can be maintained under both PTQ and QAT. A key contribution is the proposed $\lambda$-ShiftGELU activation, which addresses the limitations of uniform quantization in handling long-tailed activation distributions and proved essential in stabilizing both calibration and training. Our ablation study further confirmed that $\lambda$-ShiftGELU significantly improves approximation fidelity compared to ShiftGELU, directly translating into higher segmentation accuracy. We simplify the whole architecture by removing the L2 normalization layer and substituting bilinear upsampling with nearest-neighbor interpolation, improving hardware-friendliness without sacrificing segmentation quality. Our experiments with TVM confirm that I-Segmenter remains competitive in a true integer-only environment, demonstrating the viability of high-performance quantized inference. Further improvements may be achieved through custom kernels, enhanced operator fusion, and more reliable PyTorch-to-TVM translation. A promising direction is the design and implementation of integer-only upsampling strategies and L2 normalization approximations. Successfully tackling these issues could mitigate accuracy loss and fully unlock the potential of integer-only deployment.

%% file: A1_pseudocode.tex
\section{\texorpdfstring{Pseudo-code of the Proposed $\lambda$-ShiftGELU}
                         {Pseudo-code of the Proposed lambda-ShiftGELU}}

\label{app1}

\begin{algorithm}[H]
    \footnotesize  
    \tcp{This algorithm implements integer-only GELU. It first computes the
    integer exponential approximation using a lambda-adjusted version of ShiftExp, then applies the integer
    division to compute the GELU output.}
    \renewcommand{\baselinestretch}{1.1}\selectfont  
    \caption{\textbf{$\lambda$-ShiftGELU}}
    \label{alg:shiftGELU}
    \SetAlgoLined
    \vskip 0.05in
    \KwIn{
    \\
    $I_{x}$: Integer input \\
    $S_{x}$: Input scaling factor \\
    $k_{out}$: Output bit-precision \\
    $k_{inter}$: Intermediate bit-precision \\
    $\lambda$: Left bound scalar \\
    }
    
    \vskip 0.1in
    \SetKwFunction{FSHIFT}{$\lambda$-ShiftExp}
    \SetKwProg{Fn}{Function}{:}{}
    \Fn{\FSHIFT{$I, S, k_{inter}, \lambda$}}{
    $I_e \leftarrow I + (I \gg 1) - (I \gg 4)$ 
    \hfill {\color{gray}$\triangleright$ $I \cdot \log_2 e$} \\
    
    $I_0 \leftarrow \lfloor 1/S \rceil$ \\
    
    $I_e \leftarrow \texttt{clamp\_min}(I_e, \lambda \cdot k_{inter} \cdot (-I_0))$ 
    \hfill {\color{gray}$\triangleright$ Eq.~\ref{eq:lambda_shift_gelu}} \\
    
    $q \leftarrow \lfloor I_e / (-I_0) \rfloor$ 
    \hfill {\color{gray}$\triangleright$ Integer part} \\
    
    $r \leftarrow -(I_e - q \cdot (-I_0))$ 
    \hfill {\color{gray}$\triangleright$ Decimal part} \\
    
    $I_b \leftarrow ((-r) \gg 1) + I_0$ \\
    
    $I_{exp} \leftarrow I_b \ll (k_{inter} - q)$
    \hfill {\color{gray}$\triangleright$ Eq.~\ref{eq:decimal_and_integer}} \\
    
    $S_{exp} \leftarrow S / (2^{k_{inter}})$ \\
    
    \Return $(I_{exp}, S_{exp})$ 
    \hfill {\color{gray}$\triangleright$ $S_{exp}\cdot I_{exp}\approx e^{S \cdot I}$}
    }
    
    \vskip 0.1in
    \SetKwFunction{FGELU}{$\lambda$-ShiftGELU}
    \SetKwProg{Fn}{Function}{:}{}
    \Fn{\FGELU{$I_{x}, S_{x}, k_{out}, k_{inter}, \lambda$}}{
    $I_p \leftarrow I_{x} + (I_{x} \gg 1) + (I_{x} \gg 3) + (I_{x} \gg 4)$ 
    \hfill {\color{gray}$\triangleright$ $1.702I$} \\
    
    $I_{max} \leftarrow \max(I_p)$ \\
    
    $I_{\Delta} \leftarrow I_p - I_{max}$ \\
    
    $(I_{exp}, S_{exp}) \leftarrow \texttt{ShiftExp}(I_{\Delta}, S_{x}, k_{inter}, \lambda)$ \\
    
    $(I_{exp}', S_{exp}') \leftarrow \texttt{ShiftExp}(-I_{max}, S_{x}, k_{inter}, \lambda)$ \\
    
    $(I_{div}, S_{div}) \leftarrow \texttt{IntDiv}(I_{exp}, I_{exp} + I_{exp}', k_{out})$ 
    \hfill {\color{gray}$\triangleright$ Eq.~\ref{eq:IntDiv}} \\
    
    $(I_{out}, S_{out}) \leftarrow (I_{x} \cdot I_{div},\, S_{x} \cdot S_{div})$ \\
    
    \Return $(I_{out}, S_{out})$ 
    \hfill {\color{gray}$\triangleright$ $I_{out}\cdot S_{out} \approx \text{GELU}(I_{x}\cdot S_{x})$}
    }
    \vskip 0.05in
\end{algorithm}

%% file: A2_prediction_comparison.tex
\section{Additional Experimental Results}
\label{sec:vis)}

\subsection{Qualitative Segmentation Results} 
\label{sec:app2}

\begin{figure}[htbp]
    \centering

    \includegraphics[width=1.0\textwidth]{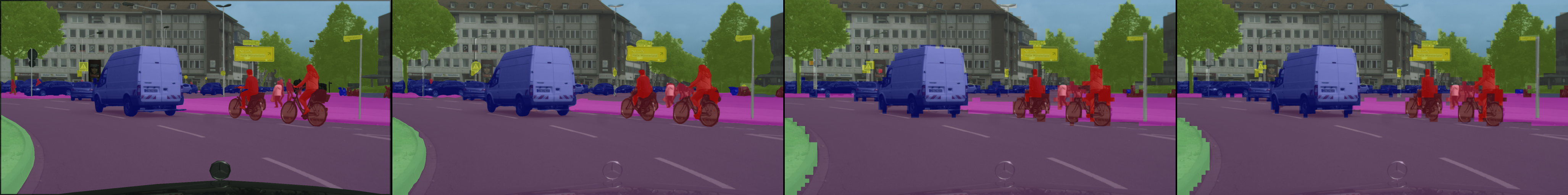}
 
    \includegraphics[width=1.0\textwidth]{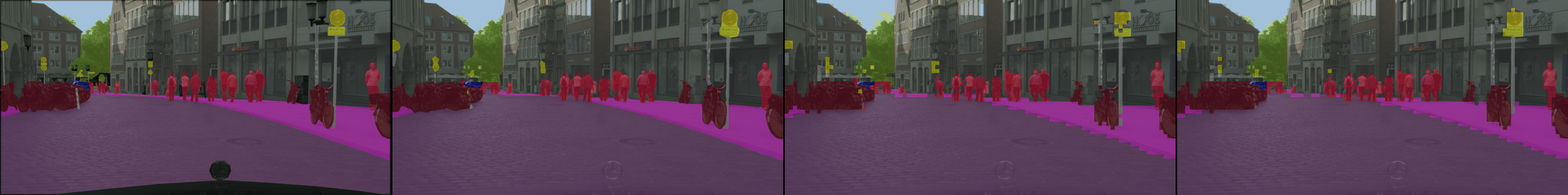}

    \includegraphics[width=1.0\textwidth]{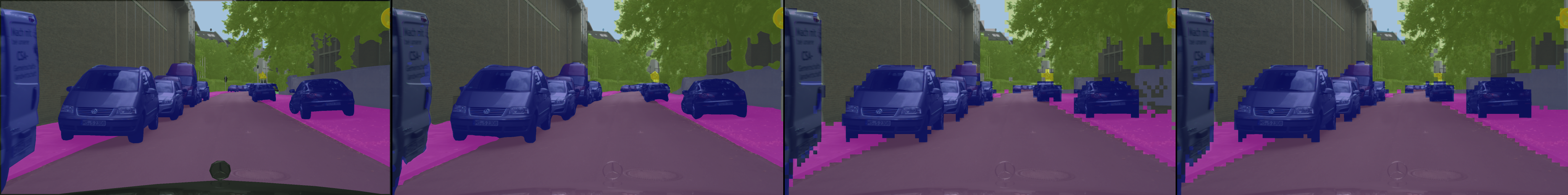}

    \includegraphics[width=1.0\textwidth]{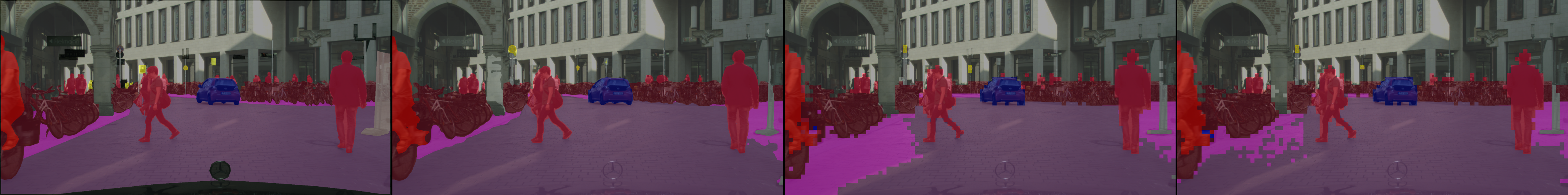}

    \includegraphics[width=1.0\textwidth]{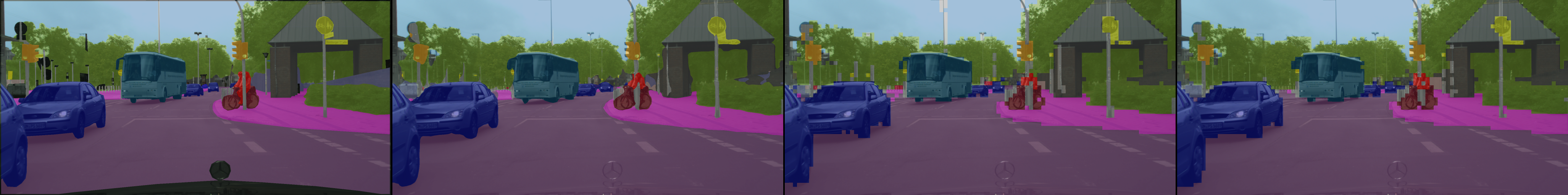}
    
    \includegraphics[width=1.0\textwidth]{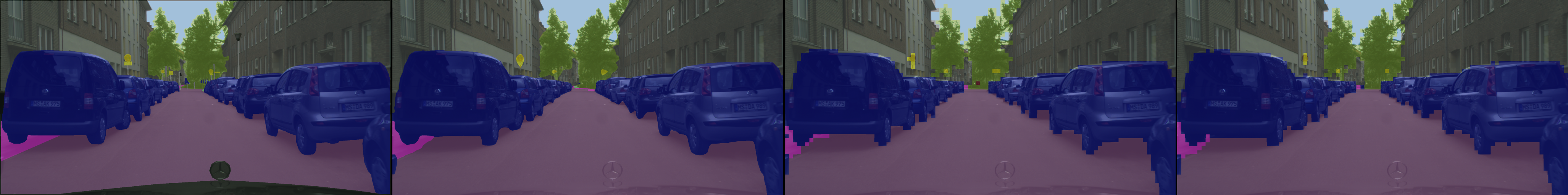}
    
    \includegraphics[width=1.0\textwidth]{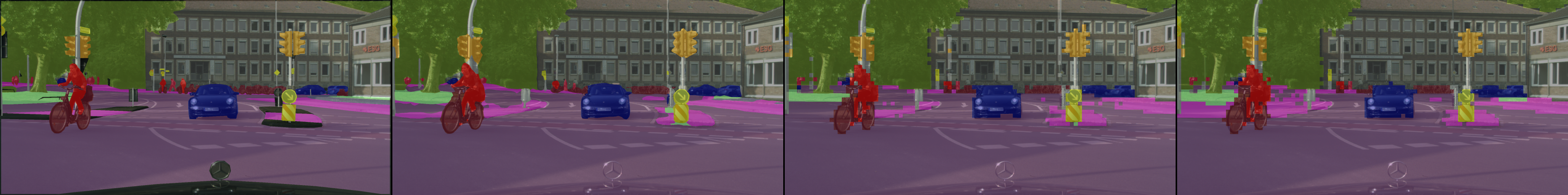}

    \caption{Qualitative segmentation results on Cityscapes. From left to right: ground truth, FP32 Segmenter, I-Segmenter with PTQ, and I-Segmenter with QAT.}
    \label{fig:appendix_predictions}
\end{figure}

\newpage
\subsection{Confusion Matrix Analysis}
\label{app3}

\begin{figure}[htbp]
    \centering

    \begin{minipage}{0.48\textwidth}
        \centering
        \includegraphics[width=\linewidth]{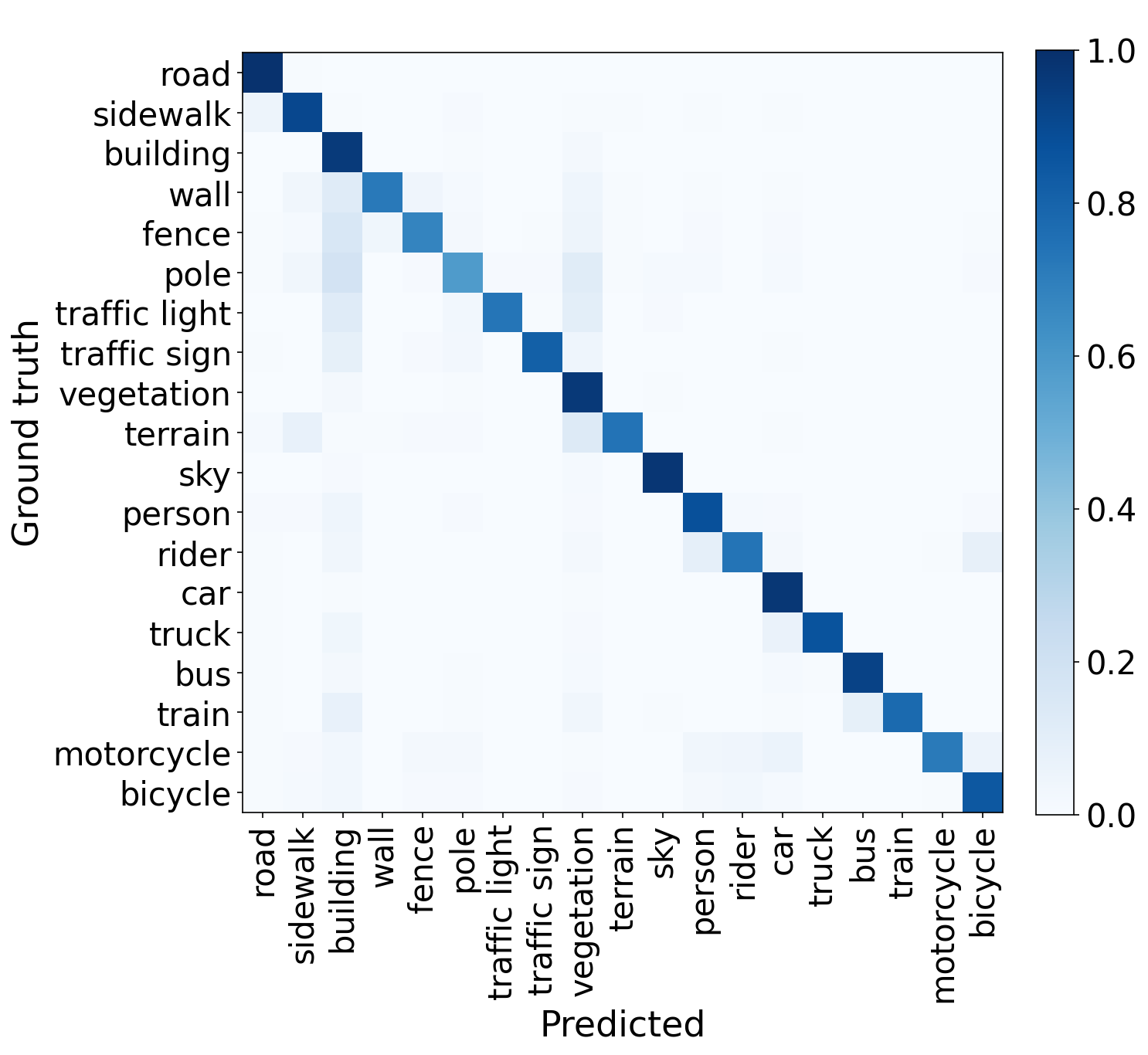}
    \end{minipage}
    \hfill
    \begin{minipage}{0.48\textwidth}
        \centering
        \includegraphics[width=\linewidth]{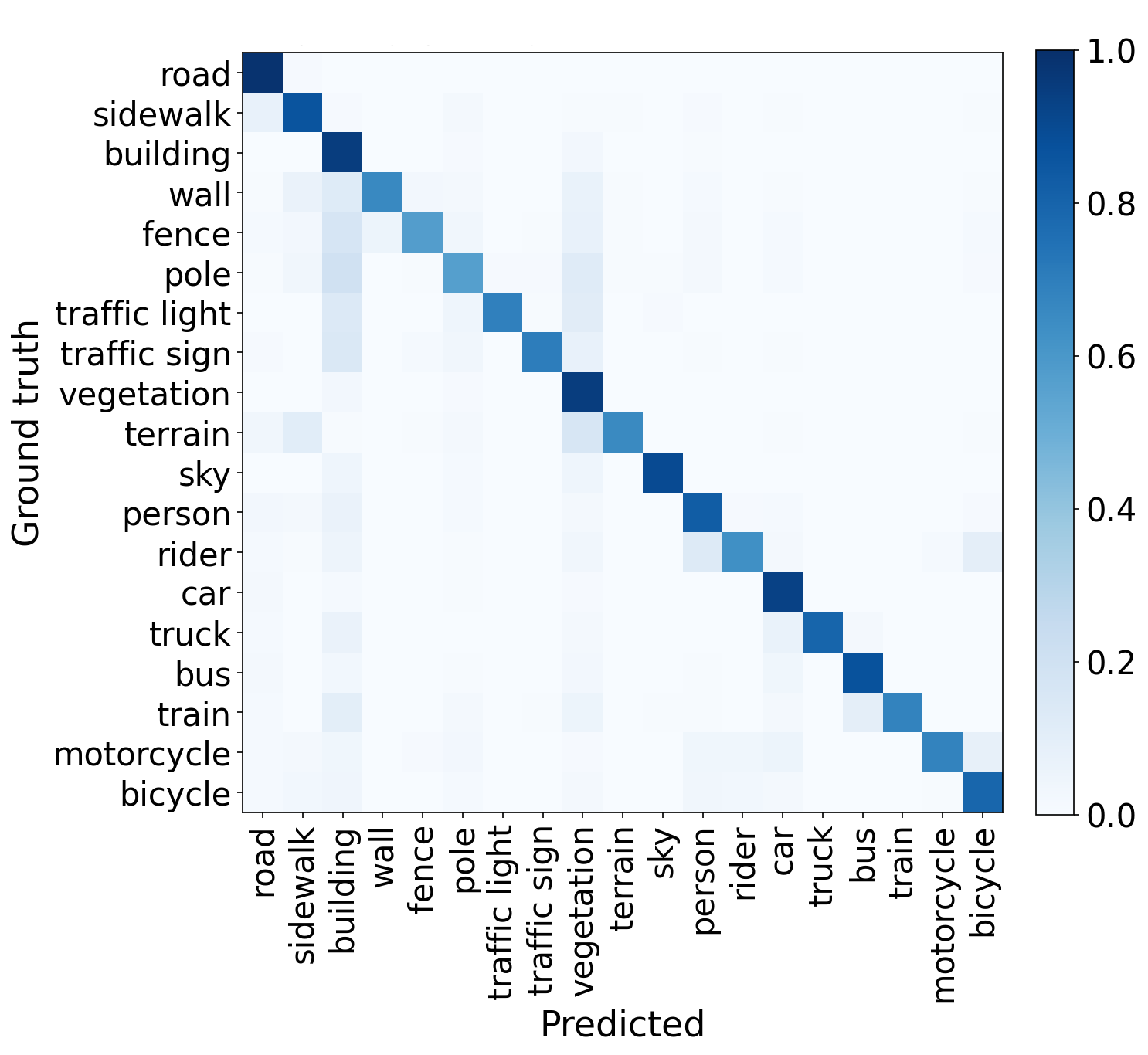}
    \end{minipage}

    \caption{Row-normalized confusion matrices on Cityscapes comparing the FP32 Segmenter (left) and the quantized I-Segmenter under PTQ with a calibration set of size 500 (right). Both models exhibit a strong diagonal structure, indicating that most classes are correctly predicted after quantization. Compared to FP32, the PTQ model shows a slight increase in inter-class confusion, particularly for visually similar categories (e.g., \textit{sidewalk}, \textit{terrain}, \textit{building}) and thin or small-scale objects (e.g., \textit{pole}, \textit{traffic sign}, \textit{traffic light}). These deviations appear as a mild dispersion of mass away from the diagonal, reflecting reduced fine-grained discrimination. In contrast, large and well-defined classes such as \textit{road}, \textit{sky}, and \textit{car} remain largely unaffected, prserving strong diagonal dominance.}
    \label{fig:confusion_matrix}
\end{figure}

\begin{figure}[htbp]
    \centering

    \begin{minipage}{0.48\textwidth}
        \centering
        \includegraphics[width=\linewidth]{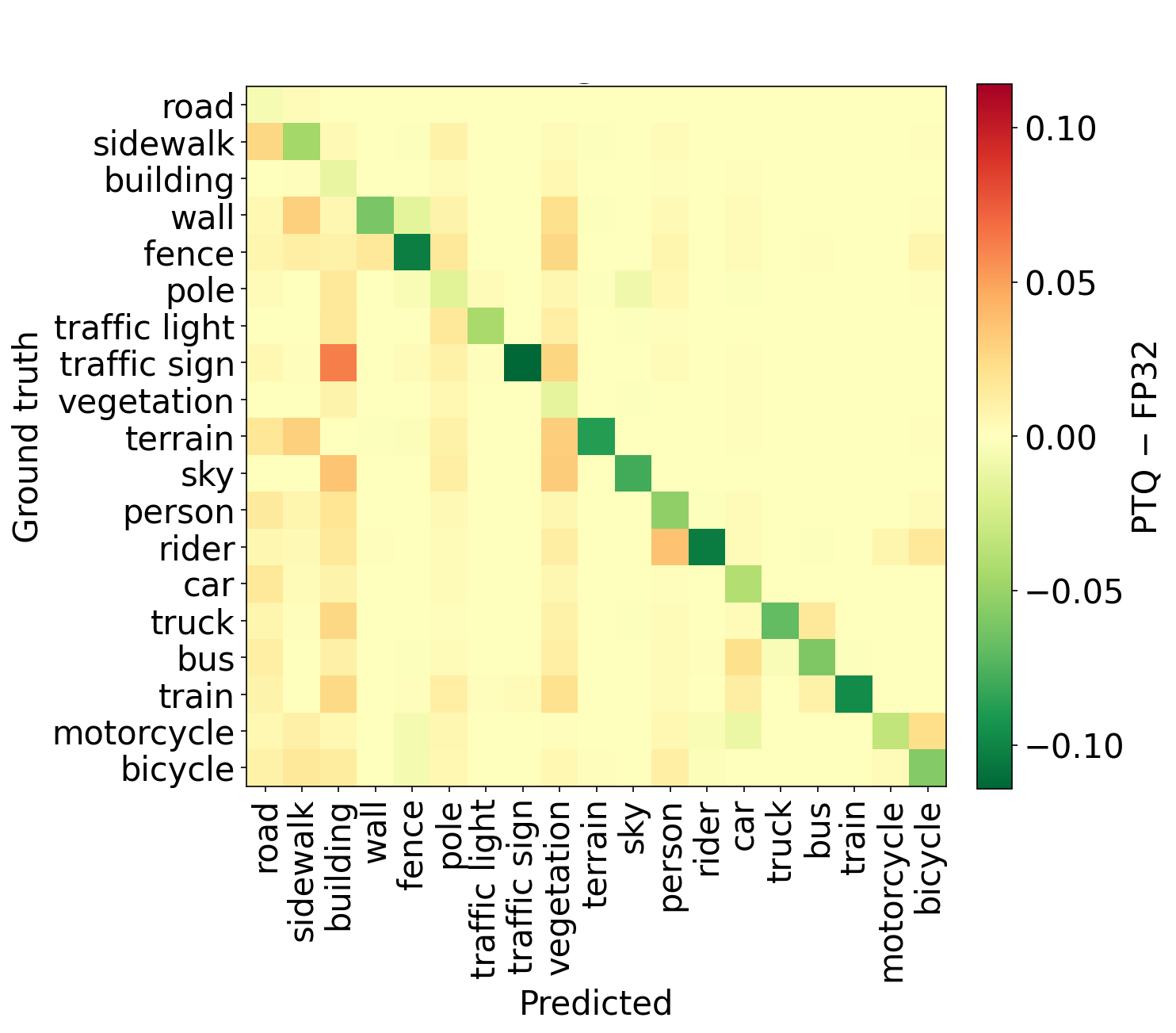}
    \end{minipage}
    \hfill
    \begin{minipage}{0.48\textwidth}
        \centering
        \includegraphics[width=\linewidth]{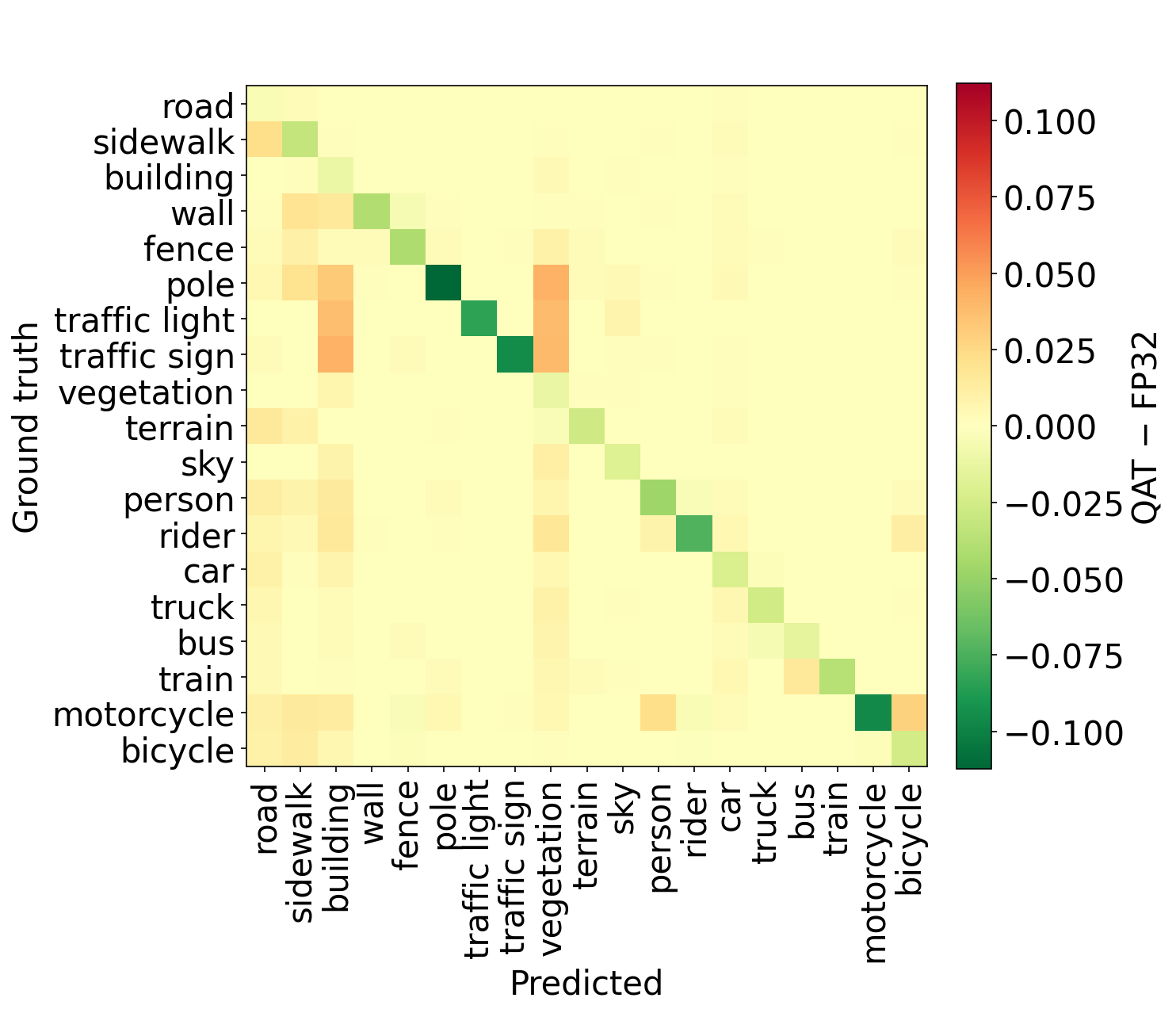}
    \end{minipage}
    \caption{Difference of confusion matrices relative to the FP32 baseline (left: PTQ, right: QAT). Color coding indicates relative performance (red: quantized model improves over FP32, green: FP32 performs better). 
    }
    \label{fig:diff_confusion_matrix}
\end{figure}


\begin{figure}[htbp]
    \centering
    \includegraphics[width=1.0\textwidth]{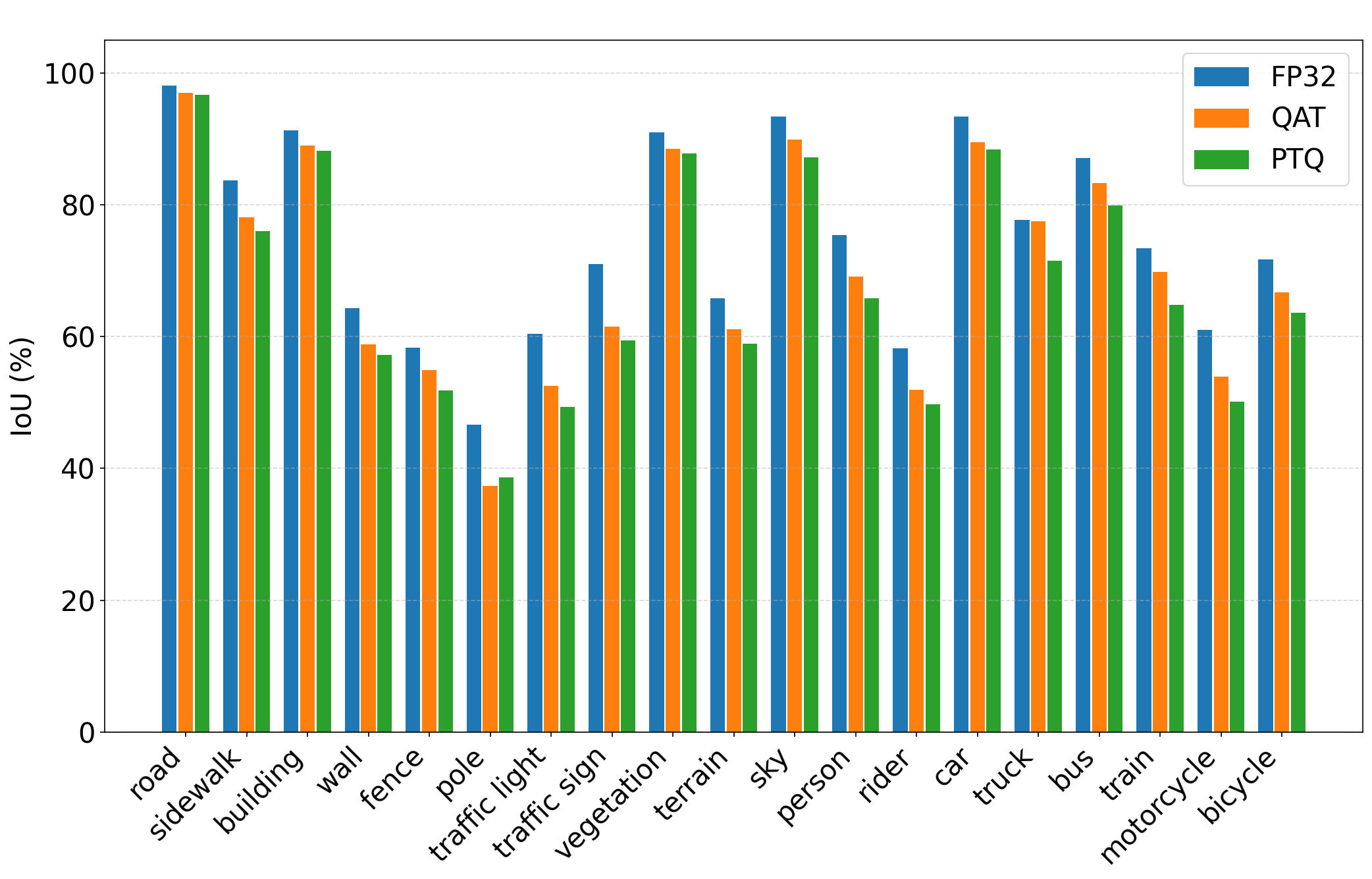}
    \caption{Per-class IoU (\%) illustrating the impact of quantization. QAT generally yields higher accuracy than PTQ across classes, except for the \textit{pole} category where PTQ performs slightly better.}
    \label{fig:Per-classIoU}
\end{figure}